\documentclass[runningheads]{llncs}
\usepackage{amsmath}
\renewcommand{\vec}[1]{\bm{#1}} % \vec{v} 输出粗体 v
\usepackage[most]{tcolorbox}  % 用于绘制带背景的框
\usepackage{enumitem}         % 用于调整列表间距
\usepackage{xcolor}           % 用于颜色定义
\usepackage{graphicx}         % 用于插图
\PassOptionsToPackage{dvipsnames,table}{xcolor}
\usepackage[table]{xcolor} 
\definecolor{darkyellow}{rgb}{0.93, 0.70, 0.13}
\usepackage{siunitx}
\usepackage{eccv}
\usepackage{eccvabbrv}
 \usepackage{multirow} 
\usepackage{graphicx}
\usepackage{booktabs}
\usepackage{hyperref}

\usepackage{fancyhdr}
\begin{document}

\title{GenVideoLens: Where LVLMs Fall Short in AI-Generated Video Detection?} 

\titlerunning{Abbreviated paper title}

\author{Yueying Zou\inst{1}\and
Pei Pei Li\inst{1}\and
Zekun Li\inst{2}\and
Xinyu Guo\inst{1}\and
Xing Cui\inst{1}\and Huaibo Huang\inst{3}\and Ran He\inst{3}}

\authorrunning{F.~Author et al.}

\institute{Beijing University of Posts and Telecommunications, Beijing, China
\\
\and
University of California, Santa Barbara, CA, USA\\
\and
Center for Research on Intelligent Perception and Computing, NLPR, Institute of Automation, Chinese Academy of Sciences, Beijing, China\\}

\maketitle

\begin{abstract}
% lzk version
% In recent years, AI-generated videos have become increasingly realistic and sophisticated. Meanwhile, Large Vision–Language Models (LVLMs) have shown strong potential for detecting such content. However, existing evaluation protocols largely treat the task as a binary classification problem and rely on coarse-grained metrics such as overall accuracy, providing limited insight into where LVLMs succeed or fail. 
% To address this limitation, we introduce \textbf{GenVideoLens}, a fine-grained benchmark for diagnosing LVLM capabilities in AI-generated video detection. We construct a dataset of 400 highly deceptive AI-generated videos and 100 real ones, annotated by experts across fifteen authenticity dimensions spanning perceptual, optical, physical, and temporal reasoning aspects, and evaluate eleven representative LVLMs.
% Our analysis reveals a pronounced dimensional imbalance: LVLMs perform well on perceptual and spatial cues but struggle with optical consistency, physical interactions, and temporal–causal reasoning. Temporal perturbation experiments further show that current LVLMs barely utilize temporal information.
% Overall, GenVideoLens and our analysis provide diagnostic insights into LVLM capabilities and offer guidance for developing more robust AI-generated video detectors.

In recent years, AI-generated videos have become increasingly realistic and sophisticated. Meanwhile, Large Vision–Language Models (LVLMs) have shown strong potential for detecting such content. However, existing evaluation protocols largely treat the task as a binary classification problem and rely on coarse-grained metrics such as overall accuracy, providing limited insight into where LVLMs succeed or fail.
To address this limitation, we introduce GenVideoLens, a fine-grained benchmark that enables dimension-wise evaluation of LVLM capabilities in AI-generated video detection. The benchmark contains 400 highly deceptive AI-generated videos and 100 real videos, annotated by experts across 15 authenticity dimensions covering perceptual, optical, physical, and temporal cues. We evaluate eleven representative LVLMs on this benchmark.
Our analysis reveals a pronounced dimensional imbalance. While LVLMs perform relatively well on perceptual cues, they struggle with optical consistency, physical interactions, and temporal–causal reasoning. Model performance also varies substantially across dimensions, with smaller open-source models sometimes outperforming stronger proprietary models on specific authenticity cues. Temporal perturbation experiments further show that current LVLMs make limited use of temporal information.
Overall, GenVideoLens provides diagnostic insights into LVLM behavior, revealing key capability gaps and offering guidance for improving future AI-generated video detection systems.

\end{abstract}

\section{Introduction}
\label{sec:intro}
\begin{figure}[t]
  \centering
  \includegraphics[width=\textwidth]{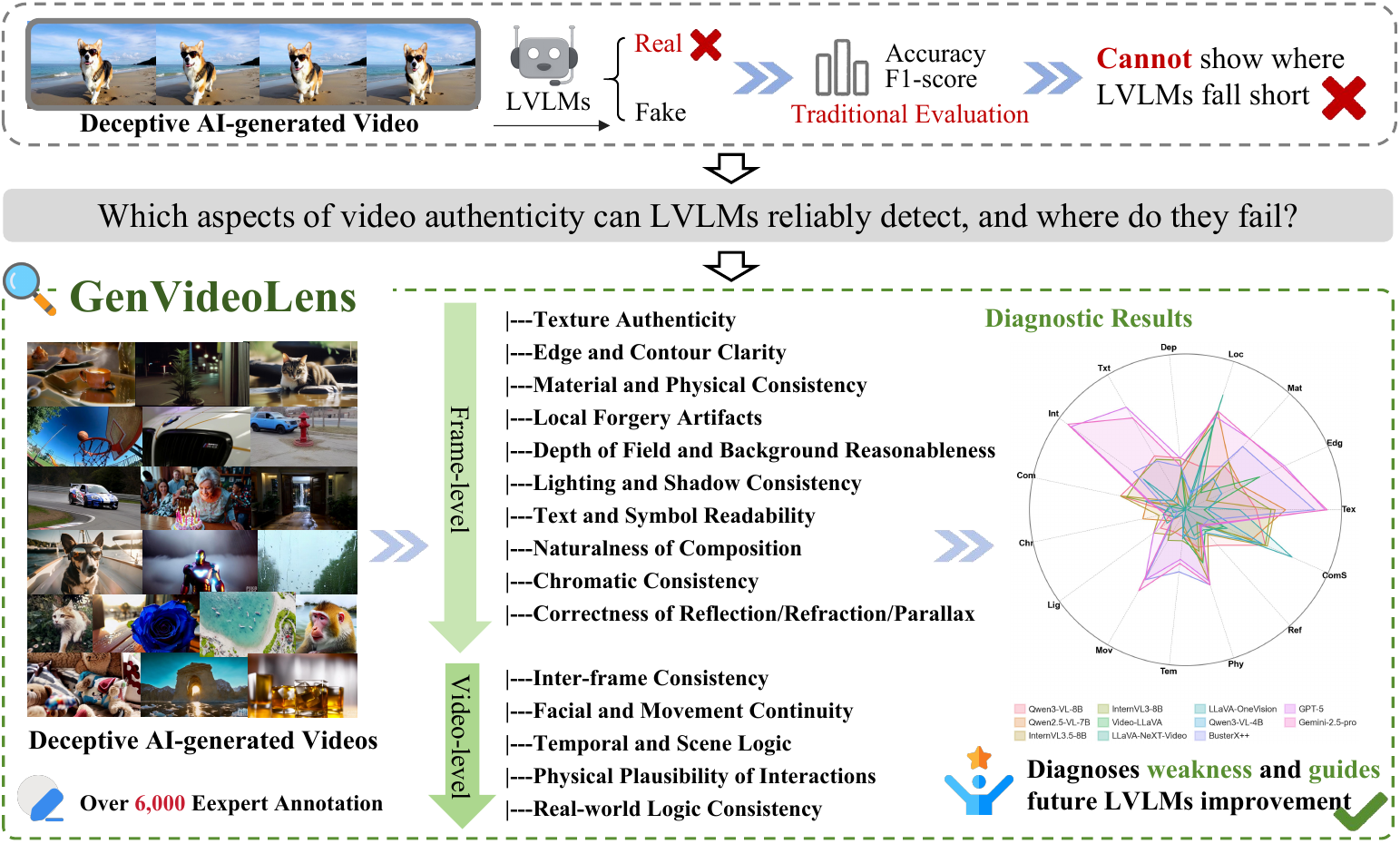}
  % \vspace{-6pt} % 减少图与caption之间距离
  \caption{\textbf{Overview of GenVideoLens.} Existing evaluations treat AI-generated video detection as a binary classification task, providing limited insight into the capabilities of LVLMs. GenVideoLens addresses this limitation by decomposing video authenticity into 15 fine-grained dimensions across frame-level and video-level analyses, enabling diagnostic evaluation that reveals where LVLMs fail and guides for future improvement.
  }
  % \vspace{-4pt} % 减少caption与下方文字的距离
  \label{fig:overview}
\end{figure}

AI-generated videos produced by modern generative models are becoming increasingly realistic and difficult to distinguish from real footage~\cite{zou2025survey, liu2024sora, zhang2025generative}. While these advances highlight the remarkable progress of video generation, they also raise serious concerns about misinformation, digital forensics, and the erosion of public trust~\cite{hwang2021effects, jin2025assessing}. To mitigate these risks, researchers have developed a variety of techniques for detecting AI-generated videos. Among them, LVLMs~\cite{gao2025david, wen2025busterx, liu2025lavid,Liu24FkaOwl,Cui25T2Agent,Liu25MMFake,zou2025survey} have recently shown strong potential for detection tasks due to their powerful visual understanding, contextual reasoning, and cross-domain generalization capabilities.

Despite these advances, understanding how well LVLMs detect AI-generated videos remains challenging. Existing benchmarks for AI-generated video detection~\cite{ni2025genvidbench, chen2024demamba, lin2025brokenvideos,Liu25VideoSafety,Xia25AVFake,Liu25MMFake} primarily frame the task as a binary classification problem and evaluate models using coarse-grained metrics such as accuracy, F1 score, or AUC. While these metrics provide an overall measure of performance, they offer limited insight into the specific capabilities that models rely on when distinguishing real videos from synthetic ones. As a result, a fundamental question remains largely unexplored: \textbf{Which aspects of video authenticity can LVLMs reliably detect, and where do they fail?} Addressing this question requires moving beyond aggregate performance metrics toward a more fine-grained understanding of model behavior.

To this end, we introduce GenVideoLens, a fine-grained benchmark designed to diagnose LVLM capabilities in AI-generated video detection. As shown in Fig.~\ref{fig:overview}, GenVideoLens organizes video authenticity into 15 authenticity dimensions, spanning perceptual cues, optical consistency, physical interactions, and temporal–causal reasoning across both frame-level and video-level analyses. The benchmark contains 400 highly deceptive AI-generated videos and 100 real videos generated from multiple state-of-the-art video generation models, ensuring diversity and realism. Each video is annotated by multiple experts, resulting in over 6,000 fine-grained labels that enable systematic evaluation across authenticity dimensions.

% Using GenVideoLens, we conduct extensive experiments on eleven representative LVLMs. Our analysis reveals a pronounced dimensional imbalance: models perform well on perceptual and spatial cues such as texture, edges, materials, and composition, but show clear weaknesses in optical cues and video-level temporal–causal reasoning. Temporal perturbation experiments further indicate that current LVLMs make limited use of temporal information when making authenticity judgments, while physical-logic tests show that many models accept visually coherent yet physically implausible events as real. Interestingly, we also observe two performance reversals: in optical and real-world logic dimensions, smaller open-source LVLMs sometimes outperform proprietary models. This suggests that open-source models tend to rely more heavily on visually salient cues aligned with human annotations, whereas proprietary models depend more on higher-level semantic priors that may overlook subtle physical inconsistencies.

Using GenVideoLens, we conduct extensive experiments on eleven representative LVLMs. Our analysis reveals a pronounced dimensional imbalance. Models perform well on perceptual cues such as texture, edges, materials, and composition, but show clear weaknesses in optical consistency and temporal causal reasoning at the video level. 
Temporal perturbation experiments further indicate that current LVLMs make limited use of temporal information when making authenticity judgments, while physical logic tests show that many models accept visually coherent yet physically implausible events as real.
We also observe notable performance variation across dimensions. In the optical and real-world logic dimensions, smaller open-source LVLMs sometimes outperform proprietary models. One possible explanation is that open-source models may rely more on visually salient cues that align with human annotations, whereas proprietary models may place greater emphasis on higher-level semantic interpretations, which could cause subtle physical inconsistencies to be overlooked.

Our contributions can be summarized as follows:
\begin{itemize}
  \item We introduce \textbf{GenVideoLens}, a fine-grained benchmark for AI-generated video detection that organizes video authenticity into fifteen dimensions. The benchmark contains 400 highly deceptive AI-generated videos and 100 real videos, with over 6,000 expert annotations across these dimensions.

  \item We propose a dimension-wise evaluation framework that enables systematic analysis of LVLM capabilities beyond aggregate binary accuracy, revealing fine-grained capability differences across models.

  \item Through extensive experiments on eleven LVLMs, we uncover key limitations of current models, including weaknesses in optical reasoning, physical interactions, and temporal–causal understanding, as well as substantial variation across models under different authenticity dimensions.
\end{itemize}

\section{Related Works}
\label{sec:formatting}

\begin{figure*}[t]
  \centering
  \includegraphics[width=\textwidth]{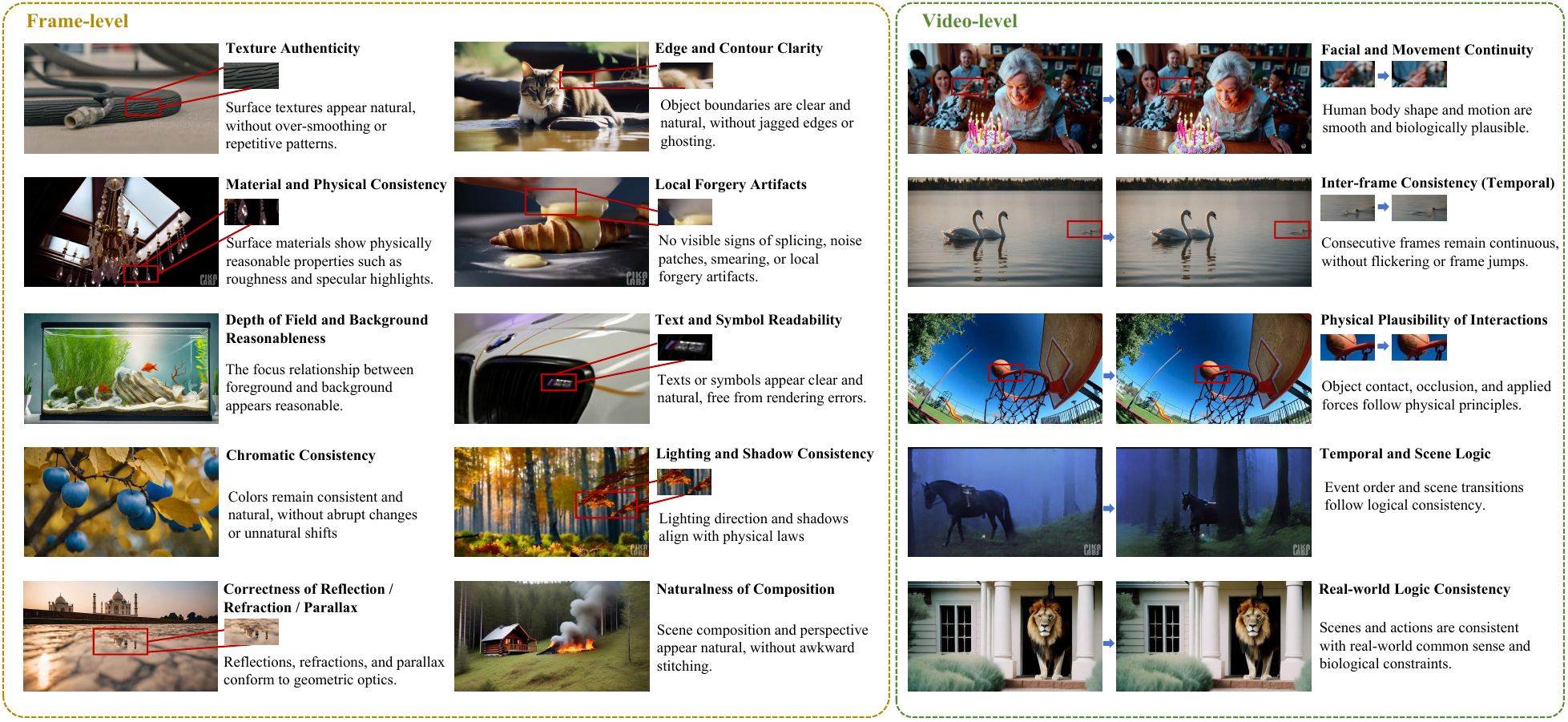}
  \caption{\textbf{Visualization of the 15 authenticity dimensions in GenVideoLens.}
    The dimensions are organized into \textit{frame-level} and \textit{video-level} categories, each illustrated with representative visual examples from the dataset to highlight the corresponding authenticity cues.}
  \label{fig:dimension_overview}
\end{figure*}

%-------------------------------------------------------------------------
\subsection{LVLMs for AI-Generated Video Detection}
AI-generated video detection has evolved rapidly. Early approaches formulated the task as a binary classification problem targeting low-level artifacts or spatiotemporal inconsistencies, such as F3Net~\cite{wei2020f3net}, \cite{vahdati2024beyond}, UNITE~\cite{kundu2025towards}, \cite{corvi2025seeing}, D3~\cite{zheng2025d3}, and \cite{liu2024turns}. With the emergence of LVLMs, recent studies have begun to leverage their semantic understanding and reasoning capabilities for more interpretable AIGV detection. Works such as MM-Det~\cite{song2024learning}, Ivy-Fake~\cite{zhang2025ivy}, and AvatarShield~\cite{xu2025avatarshield} align visual encoders with large language models to detect AI-generated videos, while subsequent research introduces structured reasoning and reinforcement learning to improve interpretability. For example, DAVID-XR1~\cite{gao2025david} couples defect-level spatiotemporal annotation with visual chain-of-thought reasoning to localize forgery cues, although its reasoning scope is limited to six prompt-engineered dimensions. BusterX~\cite{wen2025busterx} and BusterX++~\cite{wen2025busterx++} further improve detection performance via supervised fine-tuning and GRPO~\cite{guo2025deepseek}, while LAVID~\cite{liu2025lavid} introduces tool-use augmentation that enables adaptive invocation of external reasoning modules. Despite these advances, existing LVLM-based approaches still lack fine-grained, dimension-wise evaluation frameworks that reveal which aspects of video authenticity remain challenging for LVLMs, making targeted performance improvement difficult.
% AI-generated video detection has evolved rapidly. Early approaches formulated it as a binary classification task targeting low-level artifacts or spatiotemporal inconsistencies, like F3Net~\cite{wei2020f3net}, \cite{vahdati2024beyond}, UNITE~\cite{kundu2025towards}, \cite{corvi2025seeing}, D3~\cite{zheng2025d3}, \cite{liu2024turns}. Later on, with the rise of LVLMs, researchers began exploiting their semantic understanding and reasoning capabilities for AIGV interpretable detection. Works such as MM-Det~\cite{song2024learning}, Ivy-Fake~\cite{zhang2025ivy} and AvatarShield~\cite{xu2025avatarshield} align visual encoders with LLM to detect AI-generated video, while subsequent studies introduced structured reasoning and reinforcement learning to enhance interpretability. 
% For instance, DAVID-XR1~\cite{gao2025david} couples defect-level spatiotemporal annotation with visual chain-of-thought reasoning, explicitly localizing forgery cues but limiting reasoning to six prompt-engineered dimensions. BusterX~\cite{wen2025busterx} and BusterX++~\cite{wen2025busterx++} further refine accuracy via SFT and GRPO~\cite{guo2025deepseek}, whereas LAVID~\cite{liu2025lavid} demonstrates tool-use augmentation, allowing adaptive invocation of external reasoning tools. However, existing LVLM-based approaches still lack fine-grained, dimension-wise evaluation to reveal which dimensions LVLMs underperform in, making targeted performance improvement difficult.

%-------------------------------------------------------------------------
\subsection{AI-Generated Video Detection Benchmarks}
The evaluation of AI-generated content (AIGC) has also evolved rapidly, yet many existing benchmarks remain focused on static or coarse-grained modalities. Early efforts such as Forensics-Bench~\cite{wang2025forensics}, FakeBench~\cite{li2025fakebench}, and MMFakeBench~\cite{Liu25MMFake} \\primarily target text and image forgery detection. More recent datasets, including GenVidBench~\cite{ni2025genvidbench}, GenVideo~\cite{chen2024demamba}, GenBuster-200K~\cite{wen2025busterx}, and GenBuster++~\cite{wen2025busterx++}, extend AIGC detection to videos by incorporating diverse generative sources and diffusion architectures. However, these benchmarks typically emphasize source diversity rather than evaluating detection difficulty from perceptual or reasoning perspectives. AEGIS~\cite{li2025aegis} introduces a challenging subset of hyper-realistic AI-generated videos, moving toward more difficult evaluation settings. Nevertheless, the construction of the hard-test split relies solely on the judgment of a single model (Qwen2.5-VL), which may be somewhat arbitrary and could limit the objectivity and generalizability of the evaluation. In contrast, GenVideoLens selects highly deceptive videos and provides a diagnostic evaluation framework organized around 15 authenticity dimensions, enabling fine-grained analysis of LVLM capabilities across perceptual and reasoning aspects.

\section{GenVideoLens}

\subsection{15-Dimensional Evaluation Framework}

To systematically examine where LVLMs fall short in AI-generated video detection, we develop a 15-dimensional evaluation framework that forms the foundation of \textbf{GenVideoLens}. This framework organizes video authenticity cues into 15 authenticity dimensions, ranging from fine-grained perceptual cues to higher-level temporal and physical reasoning, providing a human-aligned, multi-perspective diagnostic lens into LVLM capabilities.

\begin{figure}[t]
  \centering
  \includegraphics[width=\columnwidth]{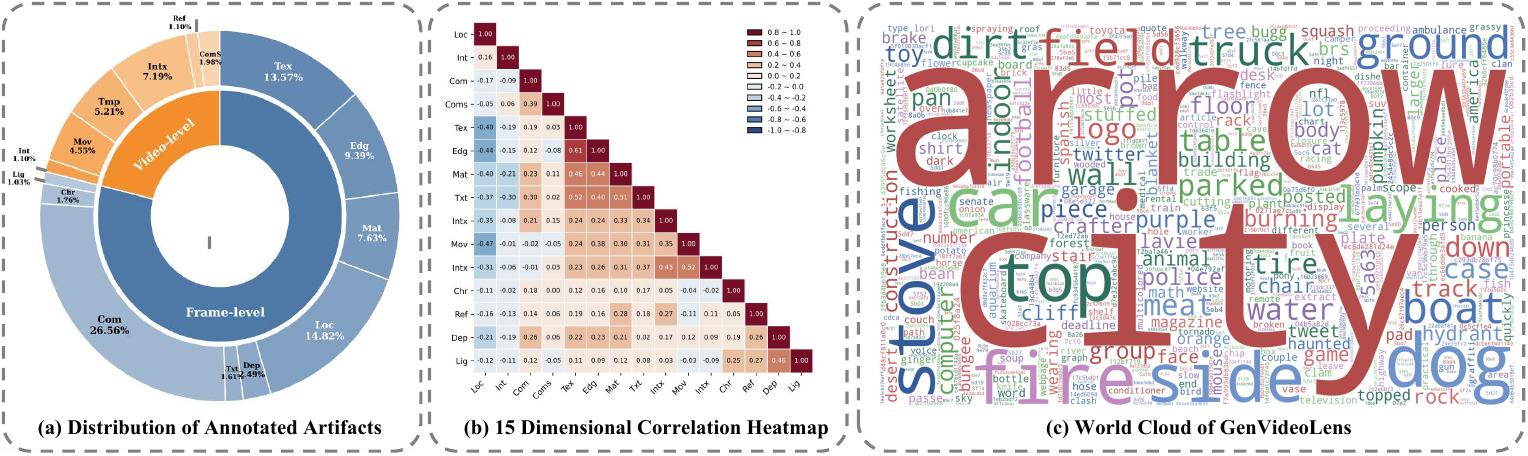}
  \caption{\textbf{Dataset statistics of GenVideoLens.} (a) Distribution of annotated artifacts across frame-level and video-level authenticity dimensions. (b) Correlation heatmap of the 15 authenticity dimensions computed from ground-truth annotations. (c) Word cloud of senmantic diversity in GenVideoLens}
  \label{fig:correlation}
\end{figure}

The design of the 15 dimensions is inspired by studies on human visual cognition~\cite{palmer1999vision, kersten2003bayesian} and physical world reasoning~\cite{battaglia2013simulation, lake2017building}. Fig.~\ref{fig:dimension_overview} illustrates the full set of dimensions together with representative visual examples.
Fig.~\ref{fig:correlation}(a) shows the distribution of annotated artifacts across the 15 dimensions. Fig.~\ref{fig:correlation}(b) presents the ground-truth correlations among these dimensions, indicating that they are largely distinguishable and exhibit limited redundancy. This observation supports the necessity of fine-grained dimension-wise evaluation rather than relying on a single aggregated metric. Fig.~\ref{fig:correlation}(c) further illustrates the semantic diversity of the dataset through a word cloud visualization.

To reflect the types of evidence required for authenticity judgment, we organize these dimensions into two complementary levels: \textit{frame-level} cues that can be assessed from individual frames, and \textit{video-level} cues that require temporal or causal reasoning across frames.

\begin{itemize}

\item \textbf{Frame-level:} focuses on static perceptual cues observable within a single frame, including texture authenticity, edge and contour clarity, material appearance consistency, local forgery artifacts (e.g., splicing, misalignment, smearing, or noise patches), depth-of-field and background reasonableness, text and symbol readability, naturalness of composition, chromatic consistency, and lighting and shadow consistency;

\item \textbf{Video-level:} emphasizes temporal and causal reasoning across frames, including inter-frame consistency and repetition, facial and movement continuity, physical plausibility of interactions (e.g., contact, occlusion, or forces), correctness of reflection, refraction, and parallax, and real-world logic consistency.

\end{itemize}

\subsection{Dataset Collection and Annotation}

\subsubsection{Data Collection.}
Our dataset consists of real videos and generated videos.

\smallskip
\noindent\textit{Real Video Collection.}
To ensure authenticity, diversity, and visual complexity, we collected approximately 8,000 candidate real videos from two complementary sources.
(1) LOKI~\cite{ye2024loki} (real subset), which contains both vertical short-form and horizontal long-form videos covering diverse real-world scenes such as humans, natural landscapes, organisms, and abiotic environments.
(2) OpenVid-1M~\cite{nan2024openvid}, a large-scale open-domain video corpus containing a wide range of semantic contexts including daily activities, urban scenes, natural environments, and social interactions.
% Together, these sources provide diverse real-world content with rich motion dynamics, illumination variations, and scene complexity.

\smallskip
\noindent\textit{Generated Video Collection.}
To cover diverse generative scenarios and mainstream video generation models, we collected approximately 11,000 AI-generated video clips as candidate synthetic samples.
All videos were sourced from three publicly available datasets:
(1) GenVideo~\cite{chen2024demamba}, which contains videos generated by multiple state-of-the-art video generation models with diverse visual styles and semantic contexts;
(2) GenVidBench~\cite{ni2025genvidbench}, which integrates outputs from several generative frameworks emphasizing temporal coherence and motion dynamics;
(3) LOKI~\cite{ye2024loki} (synthetic subset), which includes high-fidelity videos generated by recent models such as Sora~\cite{openai2024sora} and Open-Sora~\cite{opensora}.
To ensure fair and consistent evaluation, all videos were standardized through preprocessing: each clip was trimmed to 1–5 seconds, audio was removed, original aspect ratios were preserved, and resolutions were constrained to 360p–1080p.

\begin{figure}[t]
  \centering
  \includegraphics[width=\columnwidth]{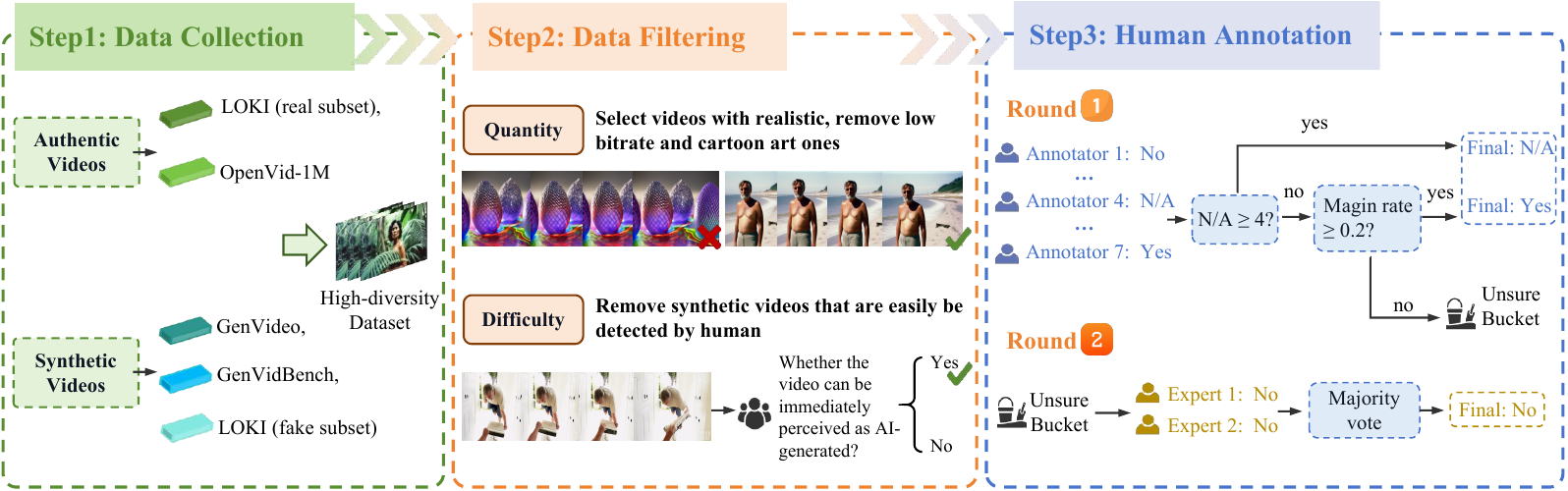}
  \caption{\textbf{Overview of the GenVideoLens dataset construction and annotation pipeline.} The process consists of data collection, dataset filtering, and human annotation.}
  \label{fig:datasets_generate}
\end{figure}

\subsubsection{Data Filtering.}
To construct a high-quality and challenging benchmark, we applied a multi-stage filtering and quality control process to both real and synthetic videos. Three volunteers with backgrounds in visual analysis manually evaluated candidate videos according to three criteria: authenticity, diversity, and perceptual difficulty.

For the real video subset, we removed samples with low visual quality, heavy occlusion, frequent scene cuts, or post-production effects, retaining only videos with natural appearance and coherent scenes. After filtering, 100 high-quality real videos were preserved, covering diverse scene types, lighting conditions, and motion patterns.

For the synthetic subset, we retained videos exhibiting high optical realism and intra-video visual coherence while discarding abstract or heavily distorted content. The remaining synthetic videos are visually convincing and difficult to distinguish from real ones by human, forming a challenging subset for authenticity detection. After multiple rounds of screening and consistency verification, 400 synthetic videos were retained. The overall data construction and filtering pipeline is illustrated in Fig.~\ref{fig:datasets_generate}.

\smallskip
\subsubsection{Human Annotation.}

To obtain reliable fine-grained annotations, we developed a web-based annotation platform specifically designed for multi-dimensional perceptual evaluation (details provided in the Appendix). Seven annotators who are graduate students with AI background participated in the annotation process and received standardized training on the annotation protocol and the definitions of the 15 authenticity dimensions.
Each annotator independently evaluated all 400 synthetic videos across the 15 authenticity dimensions using three response options: \textit{yes}, \textit{no}, and \textit{N/A}. After filtering \textit{N/A} cases, the annotation process resulted in 6,060 valid video–dimension labels.

As illustrated in Fig.~\ref{fig:datasets_generate}, we adopted a two-round labeling and aggregation procedure. For each video–dimension pair, if more than four annotators selected \textit{N/A}, the item was labeled as \textit{N/A}. Otherwise, the final consensus label was determined using a margin-based aggregation metric defined as
$m_i = \frac{|n_{\text{yes}} - n_{\text{no}}|}{n_{\text{total}}}$,
where $n_{\text{yes}}$ and $n_{\text{no}}$ denote the number of \textit{yes} and \textit{no} votes, respectively, and $n_{\text{total}}$ represents the total number of valid annotators.

Using a threshold $\tau = 0.20$, samples with $m_i > \tau$ were labeled according to the majority decision, while those with $m_i \leq \tau$ were placed into an \textbf{Unsure} bucket. All \textbf{Unsure} cases and samples with significant disagreement were subsequently re-evaluated by two experts to ensure label reliability and semantic consistency. Fig.~\ref{fig:correlation}(a) shows the distribution of annotated artifacts across the 15 authenticity dimensions of GenVideoLens.
\section{Experiments}

\subsection{Experimental Setup}

% \subsubsection{Metrics.}
% To evaluate LVLMs’ capability in detecting forged content across the 15 dimensions, we report the F1 score for the forged class, denoted as $\text{F}_1^{\text{forged}}$. In our annotation protocol, the response \textit{No} corresponds to the forged class. In Sec.~\ref{subsection:collapse}, to measure whether LVLMs can perform cross-dimensional reasoning when multiple dimensions are evaluated simultaneously. We introduce Collapse Rate $C$, $C = \frac{N_{\text{collapse}}}{N_{\text{total}}}$.
% Where, $N_{\text{collapse}}$ counts the samples for which the model produces identical \textit{binary decisions} across all active dimensions.
% % (i.e., dimensions not labeled as \textit{N/A}).
% Since the benchmark also includes a full-video binary classification task, we additionally report overall accuracy for completeness, in Sec.~\ref{sec:all_datasets}.

\subsubsection{Metrics.}

To evaluate LVLMs’ capability in detecting forged content across the 15 authenticity dimensions, we report the F1 score for the forged class, denoted as $\text{F}_1^{\text{forged}}$. In our annotation protocol, the response \textit{No} corresponds to the forged class.
When multiple authenticity dimensions are evaluated simultaneously within a single prompt (Sec.~\ref{subsection:collapse}), models may produce nearly identical predictions for all dimensions, indicating a failure to perform dimension-specific reasoning. To quantify this phenomenon, we introduce the \textbf{Collapse Rate} $C$, defined as $C = \frac{N_{\text{collapse}}}{N_{\text{total}}}$, where $N_{\text{collapse}}$ counts the samples for which the model outputs identical binary decisions across all active dimensions.
Since the benchmark also supports full-video binary classification task, we additionally report overall accuracy for completeness in Sec.~\ref{sec:all_datasets}.

\smallskip
\subsubsection{Models.}
We evaluate eleven mainstream LVLMs, including nine open-source models: Qwen3-VL-8B (Instruct), Qwen2.5-VL-7B (Instruct), Qwen3-VL-4B (Instruct), InternVL-3.5-8B, InternVL-3-8B, Video-LLaVA, LLaVA-NeXT-Video-7B, LLaVA-OneVision, and a forensic-specific LVLMs BusterX++~\cite{wen2025busterx++}. Additionally, we include two proprietary models: GPT-5 and Gemini-2.5-Pro, queried through their official APIs with default settings.
For fair comparison, all videos are uniformly sampled into 8 frames and fed into each model using a standardized prompting template (See Appendix.). All open-source models are run in FP16 precision on 2 NVIDIA RTX 5880 GPUs (48 GB each), with fixed random seeds and temperature set to 0 for reproducibility. Commercial models use deterministic decoding settings.

\definecolor{lightyellow}{RGB}{255,255,200}
\definecolor{darkyellow}{RGB}{255,230,100}
\begin{table}[t]
\caption{\textbf{Performance of LVLMs across 15 dimensions} on the GenVideoLens ($\text{F}_1^{\text{forged}}$). Higher values indicate better alignment with human judgments. The best and second-best results are highlighted in yellow and light yellow, respectively. Abbreviations: \textbf{Tex} (texture authenticity), \textbf{Edg} (edge and contour clarity), \textbf{Mat} (material and physical consistency), \textbf{Loc} (local forgery
artifacts), \textbf{Dep} (depth-of-field and background reasonableness), \textbf{Txt} (text and symbol readability), \textbf{Int} (inter-frame consistency and repetition), \textbf{Com} (naturalness of composition), \textbf{Chr} (chromatic consistency), \textbf{Lig} (lighting and shadow consistency), \textbf{Mov} (facial and movement continuity), \textbf{Tmp} (temporal logic), \textbf{Intx} (physical plausibility of interactions), \textbf{Ref} (correctness of reflection, refraction, and parallax), and
\textbf{ComS} (real-world logic consistency).}
\centering
\resizebox{\textwidth}{!}{%
% \begin{tabular}{lccccccccccccccc}
\begin{tabular}{lccc|ccc|cccc|ccccc}
\toprule
\textbf{Model} 
& \textbf{Tex} & \textbf{Edg} & \textbf{Mat} 
& \textbf{Loc} & \textbf{Dep} & \textbf{Txt} & \textbf{Com}
& \textbf{Chr} & \textbf{Lig} & \textbf{Ref}
& \textbf{Int} & \textbf{Mov} & \textbf{Tmp} & \textbf{Intx} & \textbf{ComS} \\
\midrule

Qwen3-VL-8B        
& 0.34 & 0.29 & 0.28 
& 0.22 & 0.25 & 0.30 & \cellcolor{darkyellow}0.32
& 0.07 & 0.15 & \cellcolor{darkyellow}0.23
& 0.27 & 0.12 & 0.13 & 0.19 & 0.42 \\

Qwen2.5-VL-7B      
& 0.48 & 0.29 & 0.31
& \cellcolor{lightyellow}0.50 & 0.18 & 0.09 & 0.19
& \cellcolor{darkyellow}0.21 & 0.09 & 0.10
& 0.04 & 0.03 & 0.10 & 0.26 & 0.35 \\

InternVL3.5-8B     
& 0.43 & 0.11 & 0.25
& 0.05 & 0.19 & 0.08 & \cellcolor{darkyellow}0.32
& \cellcolor{lightyellow}0.13 & \cellcolor{darkyellow}0.19 & \cellcolor{lightyellow}0.16
& 0.07 & 0.14 & 0.08 & \cellcolor{lightyellow}0.36 & \cellcolor{lightyellow}0.43 \\

InternVL3-8B       
& 0.41 & 0.26 & 0.31
& 0.02 & \cellcolor{lightyellow}0.24 & 0.28 & 0.29
& 0.00 & 0.08 & 0.08
& 0.27 & 0.17 & 0.15 & 0.21 & 0.42 \\

Video-LLaVA        
& 0.09 & 0.39 & 0.11
& 0.43 & 0.21 & 0.00 & 0.08
& 0.11 & 0.06 & 0.14
& 0.04 & 0.00 & 0.13 & 0.27 & 0.20 \\

LLaVA-NeXT-Video   
& 0.01 & 0.03 & 0.02
& \cellcolor{darkyellow}0.58 & 0.00 & 0.09 & 0.07
& 0.06 & 0.00 & 0.13
& 0.00 & 0.00 & 0.02 & 0.04 & 0.22 \\

LLaVA-OneVision    
& 0.26 & 0.04 & 0.07
& 0.29 & 0.00 & 0.00 & 0.07
& 0.11 & 0.11 & 0.09
& 0.25 & 0.00 & 0.14 & 0.09 & \cellcolor{darkyellow}0.56 \\

Qwen3-VL-4B        
& 0.11 & 0.19 & 0.20
& 0.09 & 0.17 & 0.00 & 0.20
& 0.07 & 0.07 & 0.14
& 0.08 & 0.03 & 0.12 & 0.15 & 0.40 \\

BusterX++\cite{wen2025busterx++}
& 0.62 & 0.45 & 0.41
& 0.08 & 0.21 & 0.27 & 0.31
& 0.11 & 0.04 & 0.07
& 0.39 & 0.30 & \cellcolor{darkyellow}0.38 & 0.11 & 0.12 \\

% Skyra\cite{wen2025busterx++}
% & 0.xx & 0.xx & 0.xx
% & 0.xx & 0.xx & 0.xx & 0.xx
% & 0.xx & 0.xx & 0.xx
% & 0.xx & 0.xx & 0.xx & 0.xx & 0.xx 
% \\

\midrule
GPT-5              
& \cellcolor{lightyellow}0.67 & \cellcolor{darkyellow}0.52 & \cellcolor{darkyellow}0.46
& 0.47 & \cellcolor{darkyellow}0.25 & \cellcolor{darkyellow}0.57 & 0.17
& 0.10 & \cellcolor{darkyellow}0.19 & 0.11
& \cellcolor{lightyellow}0.68 & \cellcolor{lightyellow}0.39 & 0.24 & \cellcolor{lightyellow}0.36 & 0.17 \\

Gemini-2.5-pro     
& \cellcolor{darkyellow}0.68 & \cellcolor{lightyellow}0.51 & \cellcolor{darkyellow}0.46
& \cellcolor{lightyellow}0.50 & 0.21 & \cellcolor{lightyellow}0.51 & 0.18
& 0.10 & 0.08 & 0.10
& \cellcolor{darkyellow}0.70 & \cellcolor{darkyellow}0.45 & \cellcolor{lightyellow}0.26 & \cellcolor{darkyellow}0.38 & 0.16 \\
\bottomrule
\end{tabular}}%
\label{table:15results}
\end{table}

\subsection{Dimension-wise Evaluation of LVLMs}
We evaluate representative LVLMs from three major open-source families (Qwen, InternVL, and LLaVA) together with two advanced proprietary models (GPT-5 and Gemini-2.5-pro). Since the fifteen authenticity dimensions correspond to specific forgery cues that primarily manifest in AI-generated videos, our dimension-wise analysis is conducted on the forged subset of \textbf{GenVideoLens}. Table~\ref{table:15results} reports the $\text{F}_1^{\text{forged}}$ scores for each dimension. The results reveal a clear performance imbalance across dimensions (Fig.~\ref{fig:overview}), indicating that current LVLMs exhibit uneven capabilities when reasoning about different authenticity cues.

\smallskip
\noindent\textbf{Texture Authenticity, Edge Clarity, and Material Consistency.}
LVLMs \textbf{struggle} to detect these fine-grained surface realism cues. Most open-source models exhibit limited sensitivity to artifacts such as over-smoothing, boundary inconsistencies, and repetitive textures, indicating weak discriminative ability for subtle surface details. Several models, particularly those from the LLaVA family, perform especially poorly, suggesting that current architectures still lack robust mechanisms for capturing texture fidelity and material realism.
In contrast, proprietary models consistently achieve stronger performance across these dimensions. This advantage may reflect stronger overall model capabilities, enabling more reliable assessment of subtle surface cues such as textures, edges, and material appearance.

% \smallskip
% \noindent\textbf{Local Forgery Artifacts, Depth-of-Field, Text and Symbol Readability, Naturalness of Composition.} These dimensions assess spatial consistency and geometric correctness. LVLMs display substantial variability and generally weak spatial reasoning. 
% \textit{Depth-of-filed} estimation is particularly challenging: most open-source models score between 0.00 and 0.24, reflecting limited understanding of scene geometry, background scaling, and depth structure. 
% For \textit{Local Forgery Artifacts}, LLaVA-NeXT-Video achieves the highest score (0.58), while Qwen2.5-VL-7B (0.50) and Video-LLaVA (0.43) also perform relatively well. Notably, LLaVA-NeXT-Video’s isolated strength does not generalize to its other spatial dimensions. It likely reflects model-specific behavior rather than genuinely stronger geometric reasoning.
% \textit{Naturalness of Composition} reveals similar trends, with InternVL3.5-8B achieving the best open-source score (0.32) and LLaVA variants falling below 0.10. Proprietary models again provide more stable results: GPT-5 and Gemini-2.5-pro maintain moderate performance across all spatial dimensions. This inconsistency highlights that spatial and structural reasoning remains an unsolved challenge for current LVLMs, even for stronger proprietary models.

\smallskip
\noindent\textbf{Local Forgery Artifacts, Depth-of-Field, Text and Symbol Readability, and Naturalness of Composition.}
LVLMs show \textbf{large variability} across these spatial and geometric dimensions, indicating limited spatial reasoning ability. In particular, \textit{Depth-of-Field} remains highly challenging, suggesting that current models struggle to capture scene geometry and depth structure.
While some models perform relatively well on \textit{Local Forgery Artifacts}, this advantage does not generalize to other spatial dimensions, indicating model-specific sensitivity rather than consistent geometric reasoning. A similar pattern appears in \textit{Naturalness of Composition}, where performance varies substantially across models.
In contrast, proprietary models tend to exhibit more stable performance across these spatial dimensions, although the overall performance remains moderate. These results suggest that spatial and structural reasoning remains a challenging problem for current LVLMs.

\smallskip
\noindent\textbf{Chromatic Consistency, Lighting Consistency, and Reflection Correctness.}
LVLMs perform \textbf{extremely poorly} on these optical realism dimensions, indicating limited ability to reason about illumination and physical consistency. Interestingly, smaller open-source models sometimes achieve slightly higher scores than proprietary models, although the overall performance remains low across all models.
Model reasoning inspection (See Appendix.) suggests a cue mismatch underlying this phenomenon. Proprietary models often attend to extremely subtle optical artifacts (e.g., mild chromatic noise or cross-frame hue variations), whereas human annotators rely more on salient cues such as inconsistent lighting direction, unnatural shadows, or missing reflections. Smaller open-source models tend to rely more directly on these visible surface cues, which align more closely with the signals emphasized in the annotations. This alignment explains their slightly higher scores, although all models remain weak on these physically grounded dimensions.

\smallskip
\noindent\textbf{Inter-frame Consistency, Facial and Movement Continuity, Temporal Logic, Physical Interaction, and Real-World Logic.}
These dimensions probe whether LVLMs understand temporal coherence and real-world causality. Overall, performance on this group is \textbf{consistently weak}, indicating limited ability to reason about dynamic events and causal relationships in videos. In particular, models perform poorly on \textit{Inter-frame Consistency}, \textit{Facial and Movement Continuity}, and \textit{Temporal Logic}, suggesting that current LVLMs struggle to capture motion continuity and cause–effect transitions across frames.
Interestingly, \textit{Real-World Logic} exhibits a different pattern: smaller open-source models sometimes outperform proprietary models. This likely reflects differences in reasoning strategies. Smaller models tend to rely on simple commonsense heuristics (e.g., fish cannot walk on land), whereas larger proprietary models emphasize visually plausible physical cues (such as lighting, reflections, or inertia) and may therefore accept semantically impossible but visually coherent events as real.

\begin{table}[t]
% \caption{Temporal sensitivity under frame-order perturbations. We compare sparse 8-frame inputs vs. dense all-frame inputs, reported as $\text{F}_1^{\text{forged}}$.}
\caption{\textbf{Temporal sensitivity under frame-order perturbations.} Model performance ($\text{F}_1^{\text{forged}}$) under different frame-order configurations. The results of using sparse 8-frame inputs and dense all-frame inputs are shown.}
\centering
\scriptsize
\renewcommand{\arraystretch}{1.0}
\setlength{\tabcolsep}{2.0pt}

\begin{tabular}{@{}lcccccc@{}}
\toprule
\textbf{$\text{F}_1^{\text{forged}}$} &
\multicolumn{3}{c}{\textbf{8 Frames}} &
\multicolumn{3}{c}{\textbf{All Frames}} \\
\cmidrule(lr){2-4}\cmidrule(lr){5-7}
& \textbf{ordered} & \textbf{random} & \textbf{repeated}
& \textbf{ordered} & \textbf{random} & \textbf{repeated} \\
\midrule

\multicolumn{7}{@{}l}{\textbf{Inter-frame Consistency}} \\
Qwen3-VL-8B       & 0.27 & 0.26 & 0.27 & 0.07 & 0.07 & 0.08 \\
InternVL3.5-8B   & 0.07 & 0.07 & 0.07 & 0.17 & 0.17 & 0.17 \\
\addlinespace[2pt]

\multicolumn{7}{@{}l}{\textbf{Facial and Movement Continuity}} \\
Qwen3-VL-8B       & 0.12 & 0.12 & 0.10 & 0.05 & 0.06 & 0.04 \\
InternVL3.5-8B   & 0.14 & 0.16 & 0.14 & 0.06 & 0.06 & 0.05 \\
\addlinespace[2pt]

\multicolumn{7}{@{}l}{\textbf{Physical Plausibility of Interactions}} \\
Qwen3-VL-8B       & 0.19 & 0.17 & 0.16 & 0.25 & 0.23 & 0.25 \\
InternVL3.5-8B   & 0.36 & 0.36 & 0.38 & 0.31 & 0.30 & 0.28 \\
\bottomrule
\end{tabular}
\label{table:temporal_sensitivity}
\end{table}

\subsection{Temporal Sensitivity Analysis}
To further investigate whether LVLMs genuinely utilize temporal information, we conduct a temporal sensitivity analysis by perturbing the frame order of video inputs. In addition to temporal order, we also examine the effect of temporal sampling density. Specifically, we compare inputs consisting of 8 frames with dense inputs containing all frames to determine whether limited temporal observations affect model performance.
We evaluate two representative open-source LVLM as Qwen3-VL-8B and InternVL3.5-8B under three temporal configurations: \textit{ordered}, \textit{random}, and \textit{repeated}. The evaluation focuses on three temporally sensitive dimensions: \textit{Inter-frame Consistency}, \textit{Facial and Movement Continuity}, and \textit{Physical Interaction}.

As shown in Table~\ref{table:temporal_sensitivity}, both models exhibit almost identical performance across the three configurations. Reordering or repeating frames produces negligible changes in the results, indicating that the models are largely insensitive to temporal structure.
Moreover, increasing the temporal density from sparse 8-frame inputs to all-frame inputs does not consistently improve performance. This suggests that current LVLMs do not effectively exploit additional temporal information even when more frames are available.
This near-invariance suggests that current LVLMs do not meaningfully exploit temporal dependencies when analyzing videos. Instead, their predictions appear to rely primarily on spatial semantics extracted from individual frames. This behavior explains the consistently poor performance observed in temporal reasoning dimensions such as \textit{Temporal Logic} and \textit{Inter-frame Consistency}.
Overall, these findings reveal a fundamental limitation of current LVLMs: despite operating on video inputs, they behave largely as frame-based visual analyzers rather than models capable of reasoning over temporal dynamics and causal event transitions.

% \subsection{Physical and Causal Reasoning Evaluation}
% This experiment aims to evaluate whether LVLMs possess physical-logic sensitivity, the ability to perform causal reasoning grounded in physical laws across consecutive frames.
% It further investigates whether the model relies on language templates or demonstrates genuine causal reasoning capabilities. We selected videos from the \textit{Where} Bench dataset, each containing clear motion or interaction events such as contact, occlusion, reflection, and object movement.

\begin{figure}[t]
  \centering
  \includegraphics[width=\columnwidth]{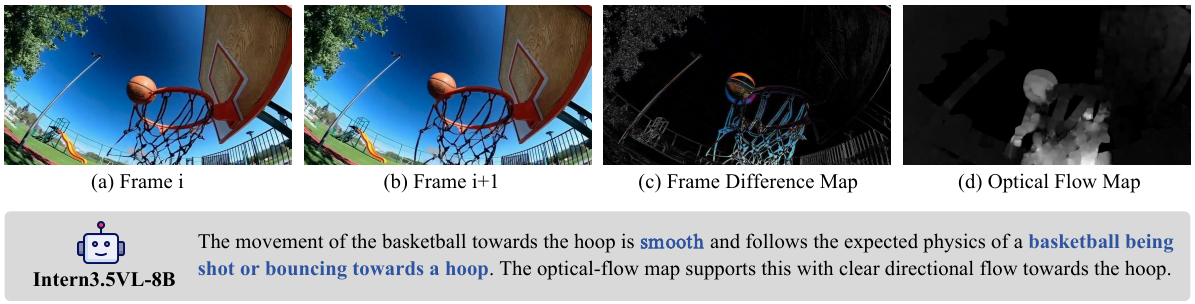}
  \caption{
  Visualization of the inputs for our physical-causal reasoning experiment.
  (a)(b) Consecutive frames showing a basketball approaching the hoop.
  (c) Frame difference map reveals the motion trajectory, and (d) optical-flow map captures directional velocity fields.
  Results shown with Intern3.5VL-8B.
  }
  \label{fig:example_diff}
  \vspace{-6pt}
\end{figure}

\begin{table}[t]
  \caption{
  Performance of Qwen3-VL-8B and InternVL3.5-8B on physical and causal reasoning. Metrics include overall accuracy (Acc$^\text{all}$), F$_1^{\text{real}}$, F$_1^{\text{forged}}$, and precision (Prec).
  }
\centering
  \small
  \setlength{\tabcolsep}{6pt}
  \renewcommand{\arraystretch}{0.85}
  \begin{tabular}{lcccc}
  \toprule
  \textbf{Model} & \textbf{Acc$^\text{all}$} & $\text{F}_1^{\text{real}}$ & $\text{F}_1^{\text{forged}}$ & $\text{Prec}$ \\
  \midrule
  Qwen3-VL-8B       & 0.51 & 0.66 & 0.11 & 0.51 \\
  InternVL3.5-8B    & 0.49 & 0.63 & 0.19 & 0.50 \\
  \bottomrule
  \end{tabular}
  \label{tab:phase3_physics}
\end{table}

\subsection{Physical and Causal Reasoning Evaluation}
This experiment evaluates whether LVLMs exhibit sensitivity to physical logic, the ability to perform causal reasoning grounded in physical laws across consecutive video frames. We further examine whether models genuinely reason about physical interactions or merely rely on superficial visual patterns. To this end, we select videos from the \textit{GenVideoLens} dataset that contain clear motion and interaction events, such as contact, occlusion, reflection, and object movement.

As observed in the previous analysis, LVLMs demonstrate limited temporal reasoning ability. To make temporal dynamics more explicit, we introduce \textit{Frame Difference Maps} and \textit{Optical Flow Maps} as auxiliary inputs~\cite{qin2025chain}. For each video, adjacent frame pairs (Frame$_i$, Frame$_{i+1}$) are used as reasoning units. Each unit consists of four visual components: Frame$_i$, Frame$_{i+1}$, \textit{Diff}, and \textit{Flow}, as illustrated in Fig.~\ref{fig:example_diff}.

Table~\ref{tab:phase3_physics} reports the performance of LVLMs on videos involving physical interactions and motion events. Despite providing motion cues through frame difference maps and optical flow maps, both models exhibit extremely low detection performance, with $\text{F}_1^{\text{forged}}$ scores of only 0.11 and 0.19. The overall accuracy remains close to random guessing (around 50\%), indicating that the models fail to reliably distinguish physically inconsistent events from authentic ones. Also, the precision of forged videos is approximately 0.5 for both models, further supporting this viewpoint.

These results indicate that current LVLMs do not effectively utilize motion cues or causal relationships across frames. Even when temporal signals such as frame differences and optical flow are provided, the models remain unable to detect violations of physical logic, such as object penetration, reflection misalignment, or abrupt acceleration changes. This suggests that current models remain largely limited to visual pattern recognition rather than true causal reasoning grounded in physical laws.

% Table~\ref{tab:phase3_physics} reports the results of the physical–logic evaluation. Although both LVLMs achieve moderate overall accuracy (0.51 for Qwen3-VL-8B and 0.49 for InternVL3.5-8B), their $\text{F}_1^{\text{forged}}$ scores for the physically implausible class remain extremely low (0.11 and 0.19, respectively).
% These results indicate that while LVLMs can perceive motion variations through frame-difference and optical-flow cues, they still fail to recognize clear physical anomalies—such as object penetration, reflection misalignment, or abrupt acceleration changes. This suggests that current models remain largely limited to visual pattern recognition rather than true causal reasoning grounded in physical laws.

\subsection{Impact of Cross-Dimension Evaluation}
\label{subsection:collapse}

\begin{figure}[t]
    \centering
    \includegraphics[width=\linewidth]{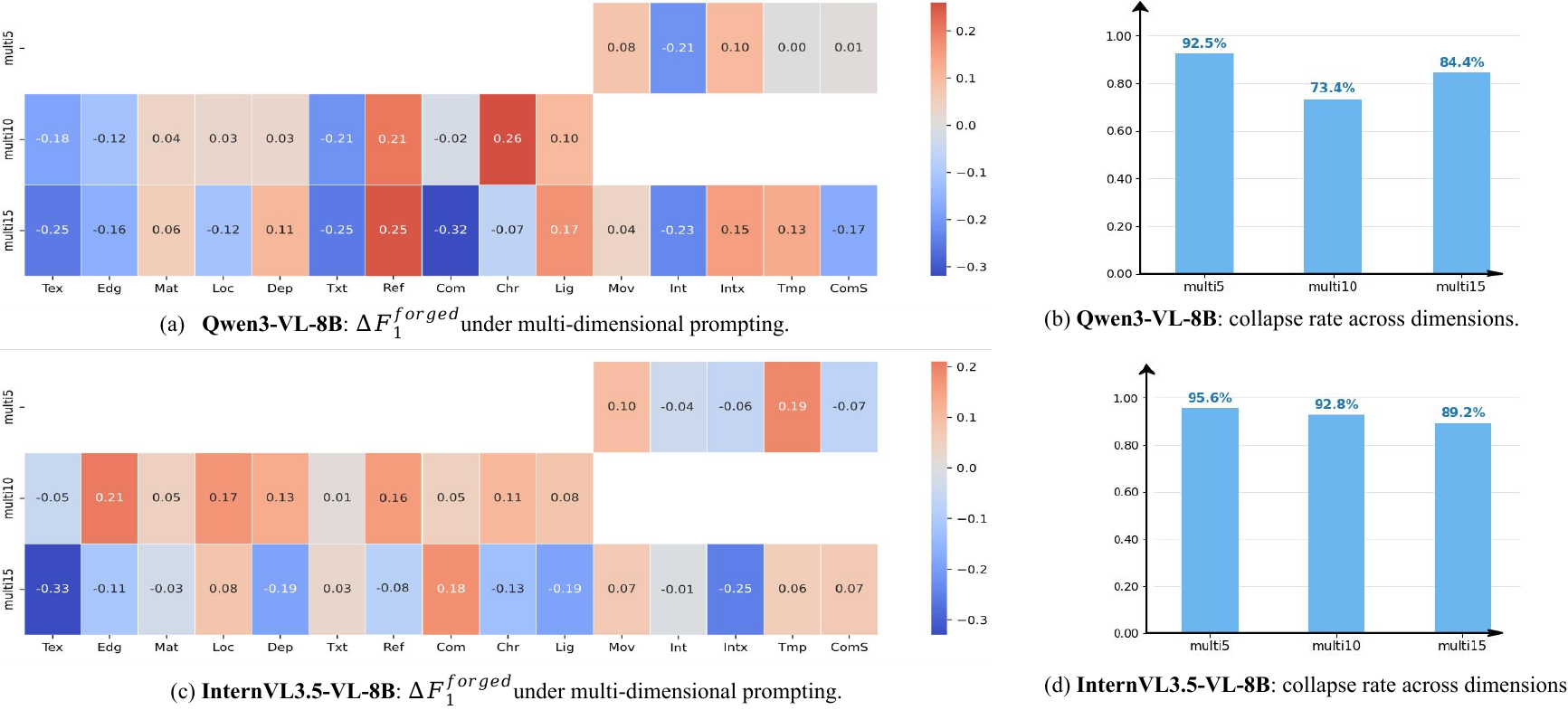}
    \caption{\textbf{Cross-Dimension Prompting Results.} Analysis of model behavior when multiple authenticity dimensions are evaluated within a single prompt. (a),(c) Heatmaps of $\Delta F_1^{\text{forged}}$ across dimensions, computed as the performance difference between multi-dimensional prompting and the default single-dimension evaluation. (b),(d) Collapse rate, measuring how often the model outputs identical binary decisions across all evaluated dimensions. The top row shows results for Qwen3-VL-8B, and the bottom row shows results for InternVL3.5-8B.}
    \label{fig:heatmap}
\end{figure}

% By default, each authenticity dimension is evaluated individually. We further investigate whether LVLMs can perform cross-dimensional reasoning when multiple authenticity dimensions are evaluated simultaneously. To examine this, we design three configurations: \textit{multi5} (five video-level dimensions), \textit{multi10} (ten frame-level dimensions), and \textit{multi15} (all dimensions) in Qwen3-VL-8B and InternVL3.5-8B. These settings progressively increase the number of authenticity dimensions presented to the model within a single prompt.

By default, each authenticity dimension is evaluated independently using a single prompt. We further examine whether LVLMs can perform cross-dimensional reasoning when multiple authenticity dimensions are evaluated simultaneously within the same prompt. To this end, we construct three prompt configurations: \textit{multi5} (five video-level dimensions), \textit{multi10} (ten frame-level dimensions), and \textit{multi15} (all dimensions). These configurations progressively increase the number of dimensions evaluated in a single prompt and are tested on Qwen3-VL-8B and InternVL3.5-8B.

As shown in Fig.~\ref{fig:heatmap}, the left heatmaps visualize the variation in $\Delta F_1^{\text{forged}}$ across dimensions, indicating how model performance changes when multiple authenticity dimensions are evaluated within a single prompt compared with the default single-dimension setting. The right bar charts report the overall collapse rate, measuring how often predictions across all evaluated dimensions converge to nearly identical outputs.

Both Qwen3-VL-8B and InternVL3.5-8B exhibit strong collapse under multi-dimensional prompts, with collapse rates exceeding 90\%. Rather than leveraging cross-dimensional cues, the models tend to produce nearly uniform predictions across dimensions. 
Qwen3-VL-8B shows large $\Delta F_1^{\text{forged}}$ fluctuations, where perceptual dimensions degrade while some reasoning dimensions slightly improve, suggesting attention shifts rather than genuine cue integration. InternVL3.5-8B collapses more conservatively with smaller $\Delta F_1^{\text{forged}}$ changes, indicating limited adaptability.
Overall, instead of enabling cross-dimensional verification, multi-dimensional inputs introduce substantial interference, preventing current LVLMs from maintaining stable and disentangled reasoning across authenticity dimensions.

% We study whether LVLMs can preserve dimension-wise reasoning when multiple authenticity dimensions are provided simultaneously. We design three configurations: \textit{multi5} (all five video-level dimensions), \textit{multi10} (all ten frame-level dimensions), and \textit{multi15} (all dimensions). 

% As shown in Fig.~\ref{fig:heatmap}, the left heatmaps visualize the $\Delta\text{F}_1(\mathrm{No})$ variation across dimensions, measuring how unstable each model becomes under multi-dimensional input. The right bar charts report the overall collapse rate, quantifying how often all dimensional outputs converge to nearly identical predictions. Both Qwen3-VL-8B and InternVL3.5-8B exhibit strong collapse under multi-dimensional prompts—their outputs become increasingly uniform, with collapse rates exceeding 90\%. Qwen3-VL-8B shows large $\Delta\text{F}_1(\mathrm{No})$ fluctuations, where perceptual dimensions degrade while some reasoning dimensions slightly improve, indicating an attention shift rather than true fusion. InternVL3.5-8B collapses more conservatively with smaller $\Delta\text{F}_1(\mathrm{No})$, reflecting limited adaptability. Overall, multi-dimensional input induces cross-dimensional interference, preventing current LVLMs from maintaining stable, disentangled reasoning.

\begin{table}[t]
\centering
\caption{
Detection accuracy on \textbf{GenVideoLens}. Comparison between Zero-shot prompting and Dimension-Guided Prompting.
Acc$^{\text{all}}$, Acc$^{\text{real}}$, and Acc$^{\text{forged}}$ denote overall, real-video, and AI-generated video accuracy.
}
\scriptsize
\sisetup{
  table-number-alignment = center,
  detect-weight = true,        % 跟随字体粗细
  detect-family = true,        % 跟随字体族
  mode = math                  % 数字以数学模式显示
}
\setlength{\tabcolsep}{3pt}
\renewcommand{\arraystretch}{0.95}

\begin{tabular}{l
S[table-format=1.2]
S[table-format=1.2]
S[table-format=1.2]
S[table-format=1.2]
S[table-format=1.2]
S[table-format=1.2]}
\toprule
\multirow{2}{*}{\textbf{Model}} &
\multicolumn{3}{c}{\textbf{Zero-shot}} &
\multicolumn{3}{c}{\textbf{Dimension-Guided}} \\
\cmidrule(lr){2-4} \cmidrule(lr){5-7}

& {Acc$^{\text{all}}$} & {Acc$^{\text{real}}$} & {Acc$^{\text{forged}}$}
& {Acc$^{\text{all}}$} & {Acc$^{\text{real}}$} & {Acc$^{\text{forged}}$} \\

\midrule
Qwen3-VL-8B          & 0.64 & 0.79 & 0.60 & \bfseries 0.65 & 0.89 & 0.60 \\
Qwen2.5-VL-7B        & 0.58 & 0.91 & 0.49 & \bfseries 0.80 & 0.00 & 1.00 \\
Intern3.5-VL-8B      & 0.42 & 0.96 & 0.28 & 0.36 & 0.94 & 0.22 \\
Video-LLaVA-7B       & 0.29 & 0.96 & 0.12 & 0.26 & 0.92 & 0.10 \\
LLaVA-Next-Video-7B  & 0.30 & 0.88 & 0.16 & \bfseries 0.79 & 0.09 & 0.96 \\
LLaVA-OneVision      & 0.62 & 0.55 & 0.64 & 0.44 & 0.85 & 0.34 \\
Qwen3-VL-4B          & 0.61 & 0.96 & 0.53 & 0.53 & 0.94 & 0.43 \\
\bottomrule
\end{tabular}

\label{table:acc_all}
\end{table}

\subsection{Impact of Dimension-Guided Prompting}
\label{sec:all_datasets}

Beyond dimension-wise evaluation, we further examine whether dimension-guided prompting improves AI-generated video detection compared with direct zero-shot prediction. Specifically, we compare two inference strategies for AI-generated video binary classification on the \textit{GenVideoLens} benchmark:
(a) \textbf{Zero-shot Prompting}, where the model directly predicts whether a video is real or AI-generated; and  
\sloppy
(b) \textbf{Dimension-Guided Prompting}, where the model is prompted to evaluate the video across the 15 authenticity dimensions before producing the final decision.
For completeness, we report binary classification accuracy on real videos, AI-generated videos, and the overall accuracy.
We evaluate seven representative open-source LVLMs: Qwen3-VL-8B, Qwen2.5-VL-7B, InternVL3.5-8B, Video-LLaVA-7B, LLaVA-Next-Video-7B, LLaVA-OneVision, and Qwen3-VL-4B.

As shown in Table~\ref{table:acc_all}, AI-generated video detection remains challenging, with most LVLMs achieving overall accuracy below 0.65. Dimension-guided prompting improves performance for some models such as Qwen3-VL-8B and Qwen2.5-VL-7B, which may indicate that explicit dimension-level analysis can help stabilize predictions. In contrast, other models such as InternVL3.5-8B and Video-LLaVA-7B show limited or even negative gains under 15-dimension prompting, suggesting that the effectiveness of structured prompting may depend on the underlying capability of the model.
Overall, these results suggest that the effectiveness of dimension-guided prompting varies across models, and global binary accuracy alone does not fully reveal the capability differences across authenticity dimensions, further motivating the need for fine-grained diagnostic evaluation.

\section{Conclusion}

In this work, we introduced \textit{GenVideoLens}, a fine-grained benchmark for diagnosing LVLM capabilities in AI-generated video detection. By organizing video authenticity into 15 dimensions, GenVideoLens enables dimension-wise evaluation that reveals capability differences across models beyond aggregate binary accuracy.
Our analysis shows a clear capability imbalance: while LVLMs perform relatively well on perceptual cues, they remain weak in optical consistency, physical interactions, and temporal–causal reasoning. We also observe substantial variation across models, where smaller open-source LVLMs sometimes outperform proprietary models on specific authenticity cues.
Further experiments including temporal perturbation, physical-logic evaluation, and multi-dimensional prompting, show that current LVLMs rely little on temporal evidence, lack stable physical constraints, and exhibit strong cross-dimensional interference. 
Overall, GenVideoLens provides diagnostic insights into the limitations of current LVLMs and offers guidance for improving temporal, physical, and multi-cue reasoning in future AI-generated video detectors.

\clearpage  % TODO FINAL: This \clearpage needs to be removed from both review and camera-ready versions.

% \section*{Acknowledgements}
% Please insert your acknowledgments here.

\bibliographystyle{splncs04}
\bibliography{main}

\appendix
% \clearpage
% % \documentclass[10pt,a4paper]{article}  % 或 letterpaper

% \setcounter{page}{1}
% \maketitlesupplementary

\section*{Appendix}
\renewcommand{\thesection}{\arabic{section}}
\renewcommand{\thesubsection}{\arabic{section}.\arabic{subsection}}
This supplementary material provides additional details of the GenVideoLens benchmark, including dataset construction, annotation protocols, prompting templates, and case study of 15-dimensional outputs.

\section{Dataset Details}
\subsection{Data sources}
To construct a challenging and diverse benchmark for AI-generated video detection, we collect candidate videos from several publicly available datasets that include both real-world and synthetic videos.
\begin{itemize}
    \item \textbf{GenVideo~\cite{chen2024demamba}.}  GenVideo contains over one million videos generated by multiple text-to-video models across diverse visual domains, which provides rich generative diversity covering different generation pipelines and scene categories, which makes it a valuable source of candidate synthetic videos for authenticity analysis. 
    \item \textbf{GenVidBench~\cite{ni2025genvidbench}.} GenVidBench contains videos generated by eight state-of-the-art video generation systems and emphasizes cross-generator diversity and semantic richness, which helps avoid bias toward specific generative models.
    \item \textbf{LOKI~\cite{ye2024loki}.} To include both real and synthetic video sources, we further utilize the LOKI dataset. The real subset provides diverse real-world videos covering scenes such as human activities, natural landscapes, animals, and urban environments. The synthetic subset contains high-fidelity AI-generated videos from recent generative models such as Sora and Open-Sora, which represent some of the most advanced video generation systems currently available.
    \item \textbf{OpenVid-1M~\cite{nan2024openvid}.} OpenVid-1M contains over one million videos collected from the web. The dataset covers a wide range of real-world scenes, including daily activities, urban environments, natural landscapes, and object interactions, providing rich semantic and motion diversity.
\end{itemize}

% \begin{table}[h]
% \centering
% \small
% \setlength{\tabcolsep}{6pt}
% \renewcommand{\arraystretch}{1.1}
% \begin{tabular}{lcc}
% \toprule
% \textbf{Generator} & \textbf{Generation Type} & \textbf{Source Dataset} \\
% \midrule
% Sora & diffusion-based video generation & LOKI \\
% Open-Sora & open-source large video model & LOKI \\
% Runway Gen-2 & commercial text-to-video system & GenVidBench \\
% Pika & text-to-video diffusion model & GenVidBench \\
% ModelScope T2V & diffusion-based generation & GenVideo \\
% VideoCrafter & latent diffusion video model & GenVideo \\
% AnimateDiff & diffusion motion adapter & GenVideo \\
% VideoLDM & latent diffusion model & GenVideo \\
% \bottomrule
% \end{tabular}
% \caption{Representative video generation models covered in
% the candidate synthetic video pool used to construct
% GenVideoLens.}
% \label{tab:generator_coverage}
% \end{table}

\subsection{Dataset Statistics}

\begin{figure}[h]
    \centering
    \includegraphics[width=0.75\linewidth]{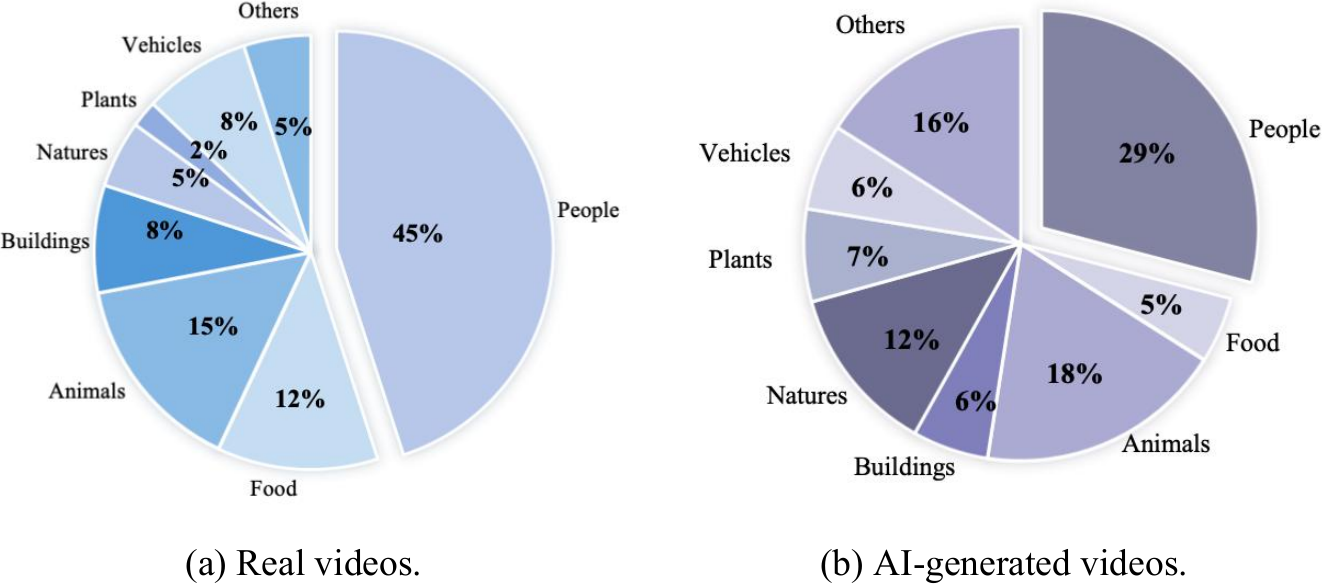}
    \caption{Scene category distribution in the GenVideoLens dataset. (a) Real videos. (b) AI-generated videos. }
    \label{fig:data_scenery}
\end{figure}

After filtering and quality control, the final GenVideoLens dataset contains 500 videos, including 100 real videos and
400 AI-generated videos. All videos are standardized during preprocessing:
\begin{itemize}
\item \textbf{Duration:} 1–5 seconds
\item \textbf{Resolution:} 360p–1080p
\item \textbf{Audio:} removed
\item \textbf{Aspect ratio:} preserved
\end{itemize}
GenVideoLens spans diverse visual domains including people, animals, buildings, natural scenes, plants, vehicles, food, and others, providing broad semantic coverage for evaluating AI-generated video detection (Fig.~\ref{fig:data_scenery}).

\subsection{Dimension Label Distribution}

Table~\ref{tab:dimension_distribution} reports the label distribution across the 15 authenticity dimensions. GenVideoLens is constructed by intentionally selecting visually convincing and high-fidelity AI-generated videos. This distribution reflects the realistic difficulty of detecting high-quality AI-generated videos, where most frames appear visually plausible while only certain authenticity cues reveal inconsistencies. \textbf{Moreover, our evaluation focuses on the forged class (\textit{No}). $\text{F}_1^{\text{forged}}$ directly measures the ability of models to detect authenticity violations, making the evaluation robust to label imbalance.}

\begin{figure}[h]
\centering
\includegraphics[width=\linewidth]{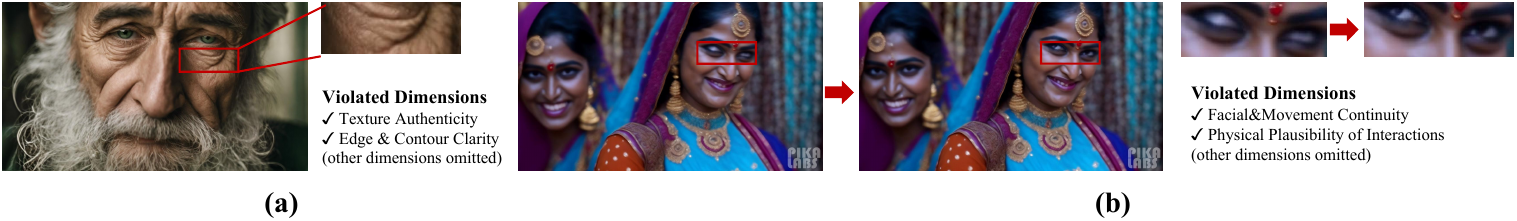}

\caption{
Examples of authenticity violations in GenVideoLens dimensions.
(a) A synthetic frame exhibiting abnormal texture patterns and inconsistent edge structures, corresponding to violations
of \textit{Texture Authenticity} and \textit{Edge \& Contour Clarity}.
(b) A video sequence showing unnatural eye motion and physically implausible interactions, violating \textit{Facial \& Movement Continuity} and \textit{Physical Plausibility of Interactions}. }
\label{fig:dimension_examples}
\end{figure}

Fig.~\ref{fig:dimension_examples} illustrates typical examples of dimension violations in high-quality AI-generated videos. Although the overall appearance appears realistic, careful inspection reveals subtle inconsistencies such as unnatural texture patterns, distorted edges, or implausible motion and interactions.

\begin{table}[h]
\centering
\caption{Label distribution across the 15 authenticity dimensions.}
\small
\begin{tabular}{lccc}
\toprule
\textbf{Dimension} & \textbf{Yes} & \textbf{No} & \textbf{N/A} \\
\midrule
Texture Authenticity & 215 & 185 & 0 \\
Edge and Contour Clarity & 272 & 128 & 0 \\
Material Consistency & 293 & 104 & 3 \\
Local Forgery Artifacts & 198 & 202 & 0 \\
Depth-of-field Reasonableness & 364 & 34 & 2 \\
Text and Symbol Readability & 38 & 22 & 340 \\
Composition Naturalness & 38 & 362 & 0 \\
Chromatic Consistency & 376 & 24 & 0 \\
Lighting and Shadow Consistency & 386 & 14 & 0 \\
Reflection / Refraction / Parallax & 251 & 15 & 134 \\
Inter-frame Consistency & 380 & 15 & 5 \\
Facial and Movement Continuity & 114 & 62 & 224 \\
Temporal Logic & 327 & 72 & 2 \\
Physical Interaction Plausibility & 277 & 98 & 25 \\
Real-world Logic Consistency & 373 & 27 & 0 \\
\bottomrule
\end{tabular}

\label{tab:dimension_distribution}
\end{table}

\section{Annotation Protocol}

\subsection{Annotation Interface}

\begin{figure}[t]
\centering
\includegraphics[width=0.75\linewidth]{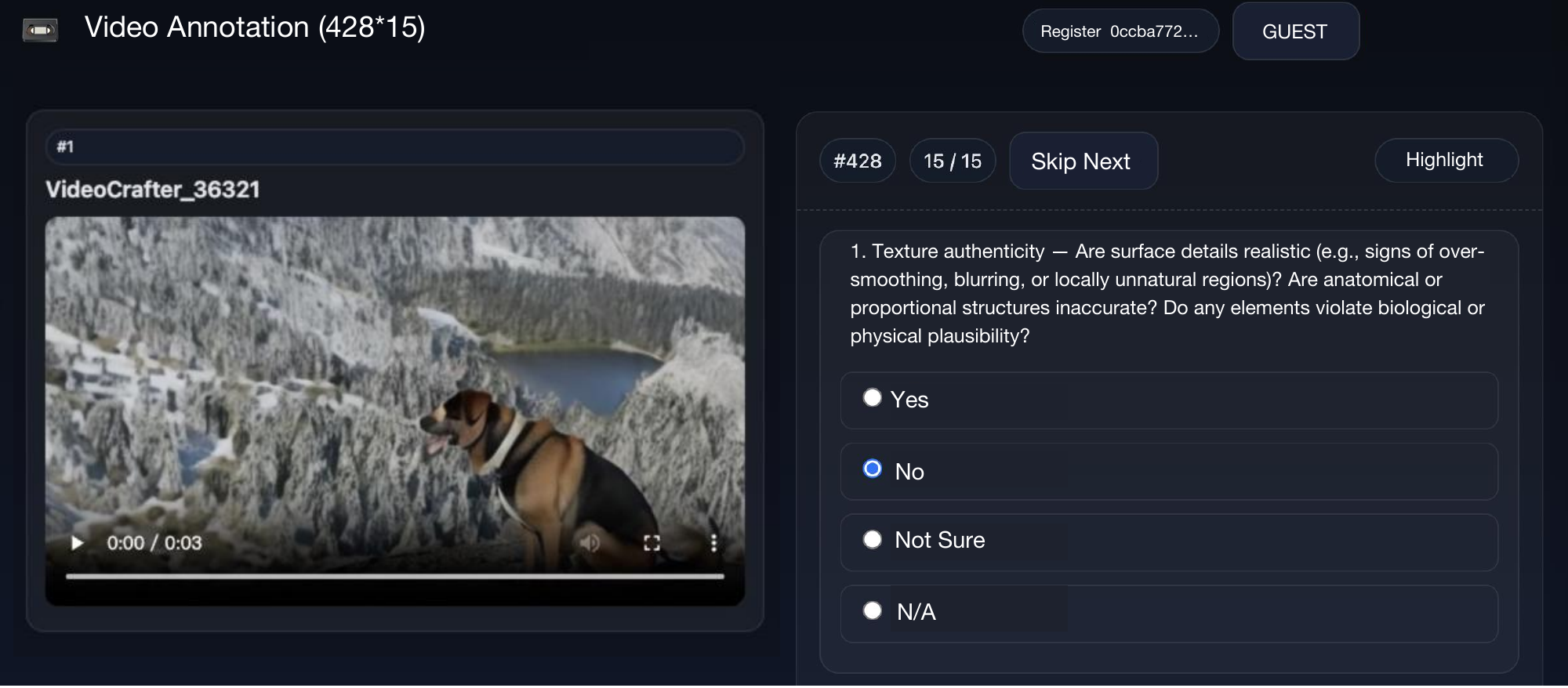}

\vspace{4pt}

\includegraphics[width=0.75\linewidth]{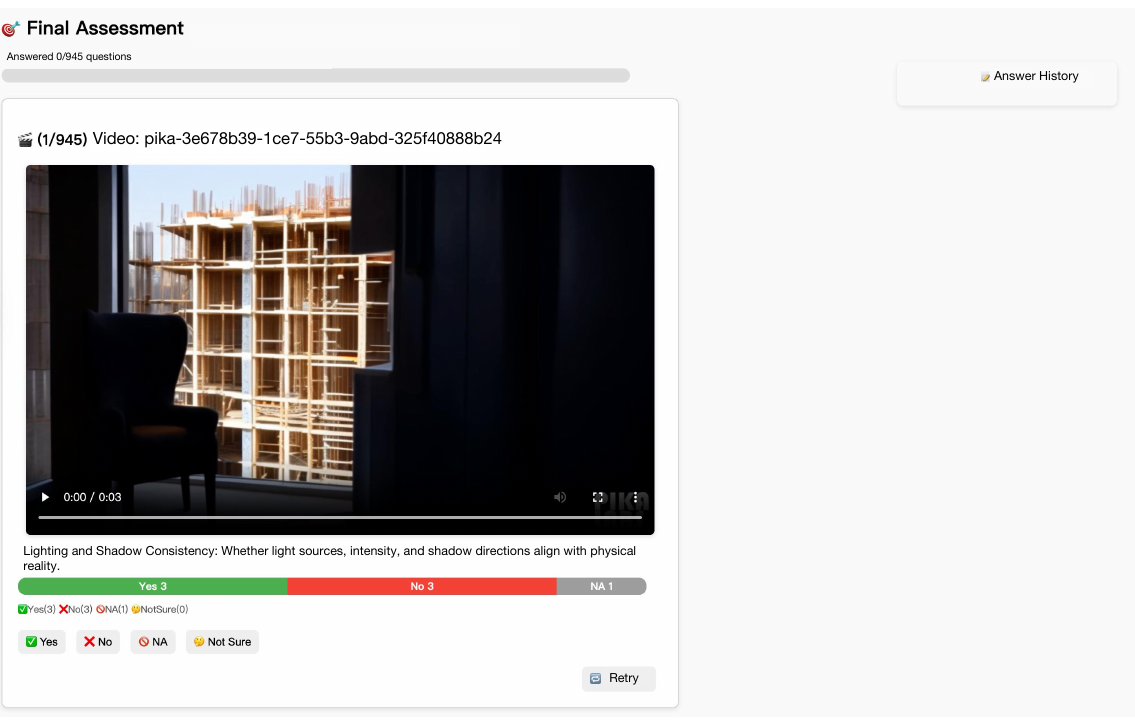}

\caption{
Annotation interface of the GenVideoLens platform.
\textbf{Top:} The main annotation website used for dimension-wise labeling.
\textbf{Bottom:} The expert review interface for conflict-bucket samples in Round-2, where reviewers can inspect previous annotation results before making the final decision.
}
\label{fig:annotation_combined}
\end{figure}

To support consistent multi-dimensional annotation, we developed a dedicated web-based annotation platform for the GenVideoLens dataset (in Fig.~\ref{fig:annotation_combined}). 
During annotation, each annotator is presented with (i) the video clip, (ii) the definition of the evaluated authenticity dimension, and (iii) three response options: \textit{Yes}, \textit{No}, and \textit{N/A}. Annotators can replay the video multiple times before submitting their decision.

The annotation process follows a two-round procedure. In the first round, annotators independently evaluate each video–dimension pair. Samples with low agreement are placed into a \textit{conflict bucket}. In the second round, expert reviewers re-evaluate these conflict-bucket samples using a dedicated review interface, where they can inspect all previous annotation results before making the final decision.

\textbf{The inter-annotator agreement measured by Fleiss'~$\kappa$ is 0.68, indicating substantial agreement and confirming the reliability of the annotations.}

\subsection{Annotator Training}

Seven annotators participated in the annotation process. All annotators were graduate students with backgrounds in artificial intelligence and computer vision. 
Before annotation, participants received a structured training session covering the definitions of the 15 authenticity dimensions, common artifacts in AI-generated videos, and representative examples from both real and synthetic videos.

Importantly, annotators were trained to evaluate each dimension independently. Different authenticity dimensions may yield conflicting judgments for the same video, and such contradictions are allowed during annotation.

\begin{figure}[t]
\centering
\includegraphics[width=0.65\linewidth]{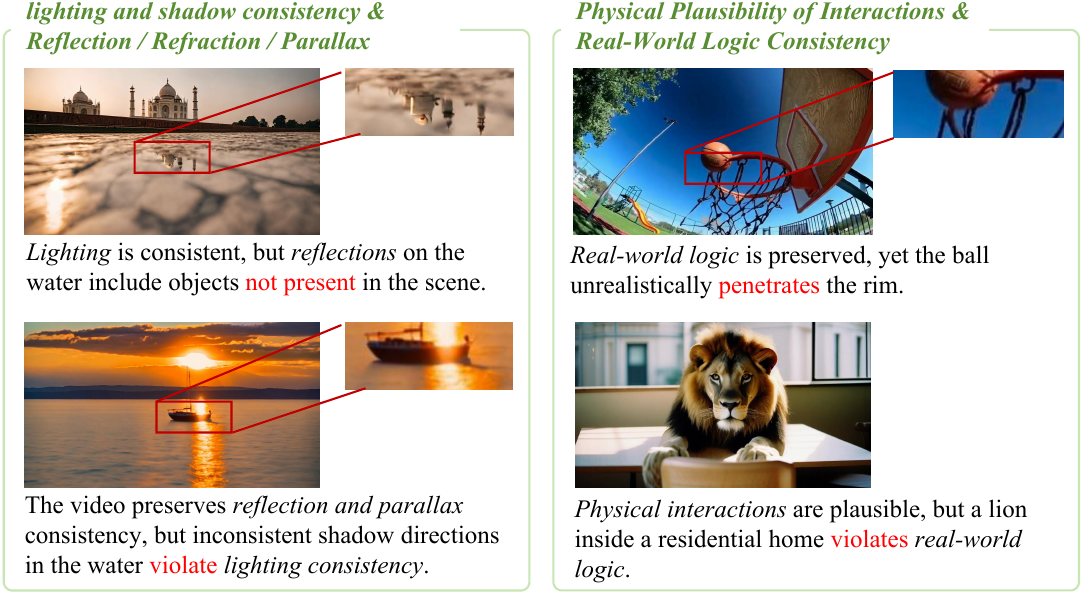}

\caption{
Examples used in annotator training to illustrate dimension-specific authenticity violations. 
Different authenticity dimensions capture distinct types of inconsistencies, enabling annotators to evaluate each dimension independently.
}
\label{fig:dimension_training_examples}
\end{figure}

Fig.~\ref{fig:dimension_training_examples} illustrates several training examples used to clarify the semantic boundaries between different authenticity dimensions. 
For example, a video may satisfy lighting consistency while violating reflection correctness, or preserve realistic physical motion while contradicting real-world logic. 
These examples help annotators understand how different dimensions capture distinct types of authenticity violations.

\subsection{Aggregation Strategy}

For each video–dimension pair, annotations were aggregated using a margin-based voting strategy. 
Let $n_{yes}$ and $n_{no}$ denote the number of positive and negative votes, respectively, and $n_{total}$ the number of valid annotators. The margin score is defined as $m_i = \frac{|n_{yes} - n_{no}|}{n_{total}}$. If $m_i > 0.2$, the majority label is used as the final annotation. Otherwise, the sample is placed into an \textit{Unsure} bucket and re-evaluated by two expert reviewers, whose consensus determines the final label.

\section{Prompt Templates}
\subsection{Prompt of 15-Dimensional Evaluation}
\label{sec:rationale}
To ensure consistent and transparent comparison across all LVLMs, we provide the full prompt used in our 15-dimensional authenticity evaluation, as shown in Fig.~\ref{fig:prompt}. The prompt explicitly instructs the model to evaluate each aspect independently, provide evidence, and produce structured outputs. This design guarantees that all models follow the same reasoning protocol and enables reproducible fine-grained analysis.

\begin{figure}[t]
\centering
\begin{tcolorbox}[
    colback=gray!15,
    colframe=black,
    boxrule=0.5pt,
    arc=0pt,
    sharp corners,
    left=6pt,
    right=6pt,
    top=6pt,
    bottom=6pt
]
\small

\noindent\textbf{Task.} Decide whether the given video is AI-generated or real-world captured. 
For every aspect, reason step-by-step and support your judgment with concrete evidence 
such as timestamps or frame numbers.

\medskip
\noindent\textbf{How to Evaluate.} Examine the aspect below. 
If the aspect does not apply (e.g., no text or characters), mark it as N/A and give a short explanation.

\medskip
\noindent\textbf{Decision Guidelines.}
\begin{itemize}[leftmargin=*, nosep, topsep=2pt, itemsep=2pt]
\item Different aspects may conflict; contradictions are allowed.
\item Provide an Overall Verdict and a Confidence Level (Low/Medium/High).
\item Always cite timestamps, frame ranges, or concrete notes as evidence.
\end{itemize}

\medskip
\noindent\textbf{Required Response Format.} Use the structure below exactly:

\medskip
\noindent\texttt{%
\begin{tabular}{@{}l@{}}
Aspect:\\
Reasoning: [What you checked]\\
Pass: Yes / No / N/A\\
Evidence: [Timestamps, frame ranges, notes]
\end{tabular}%
}

\medskip
\noindent\textbf{Notes.}
\begin{itemize}[leftmargin=*, nosep, topsep=2pt, itemsep=2pt]
\item Yes = passes check; supports real-world capture.
\item No = fails check; suggests AI generation.
\item N/A = not applicable (with a short explanation).
\end{itemize}

\medskip
\noindent\textbf{15-Dimensional Authenticity Aspects.}
\begin{enumerate}[leftmargin=*, nosep, topsep=2pt, itemsep=2pt, label=\arabic*.]
\item Texture Authenticity --- Natural surface details vs.\ oversmoothing, repetition.
\item Edge \& Contour Clarity --- Sharp boundaries vs.\ aliasing or ghosting.
\item Material \& Physical Consistency --- Plausible specularity, roughness, gloss.
\item Local Forgery Artifacts --- Splicing, misalignment, smearing, noise patches.
\item Depth-of-Field \& Background Reasonableness --- Foreground/background realism.
\item Text \& Symbol Readability --- Rendering quality, structural clarity.
\item Naturalness of Composition --- Framing, layout, perspective.
\item Chromatic Consistency --- No unnatural hue or saturation shifts.
\item Lighting \& Shadow Consistency --- Physically consistent lighting.
\item Inter-frame Consistency --- No flicker, jitter, or repeated frames.
\item Facial \& Movement Continuity --- Identity, biomechanics, natural motion.
\item Temporal \& Scene Logic --- Coherent causal ordering.
\item Physical Plausibility of Interactions --- Valid contact, occlusion, forces.
\item Reflection/Refraction/Parallax Correctness --- Consistent geometric optics.
\item Real-world Logic Consistency --- Events plausible under real-world constraints.
\end{enumerate}

\end{tcolorbox}
\caption{Full prompt used in GenVideoLens 15-dimensional authenticity evaluation.}
\label{fig:prompt}
\end{figure}

\subsection{Multi-dimension prompts}
In addition to evaluating each authenticity dimension independently, we further investigate whether LVLMs can perform cross-dimensional reasoning when multiple authenticity aspects are assessed within a single prompt. To this end, we construct two multi-dimension prompting settings based on the two-level structure of the GenVideoLens authenticity framework.
\textbf{Multi-5 (Video-level evaluation).} Evaluates the five video-level authenticity dimensions within a single prompt:
\begin{itemize}
\item Inter-frame Consistency (10)
\item Facial \& Movement Continuity (11)
\item Temporal \& Scene Logic (12)
\item Physical Plausibility of Interactions (13)
\item Real-world Logic Consistency (15)
\end{itemize}
\textbf{Multi-10 (Frame-level evaluation).} Evaluates ten frame-level authenticity dimensions within a single prompt:
\begin{itemize}
\item Texture Authenticity (1)
\item Edge \& Contour Clarity (2)
\item Material \& Physical Consistency (3)
\item Local Forgery Artifacts (4)
\item Depth-of-Field \& Background Reasonableness (5)
\item Text \& Symbol Readability (6)
\item Naturalness of Composition (7)
\item Chromatic Consistency (8)
\item Lighting \& Shadow Consistency (9)
\item Reflection/Refraction/Parallax Correctness (14)
\end{itemize}

For both settings, the prompt structure remains identical to the template described in~\ref{sec:rationale}, except that only the selected subset of dimensions is included in the \textit{Authenticity Aspects} list.

% \begin{mdframed}[backgroundcolor=gray!15, roundcorner=5pt]

% \end{enumerate}
% \end{mdframed}

% \begin{figure*}[h]
% \includegraphics[width=\textwidth]{sec/fig_appendix/prompt.pdf} 

% \label{fig:prompt} 
% \end{figure*}

\clearpage
\section{Case Study of 15-Dimensional Outputs}
\begin{figure}[ht]
  \centering
  \includegraphics[width=0.8\textwidth]{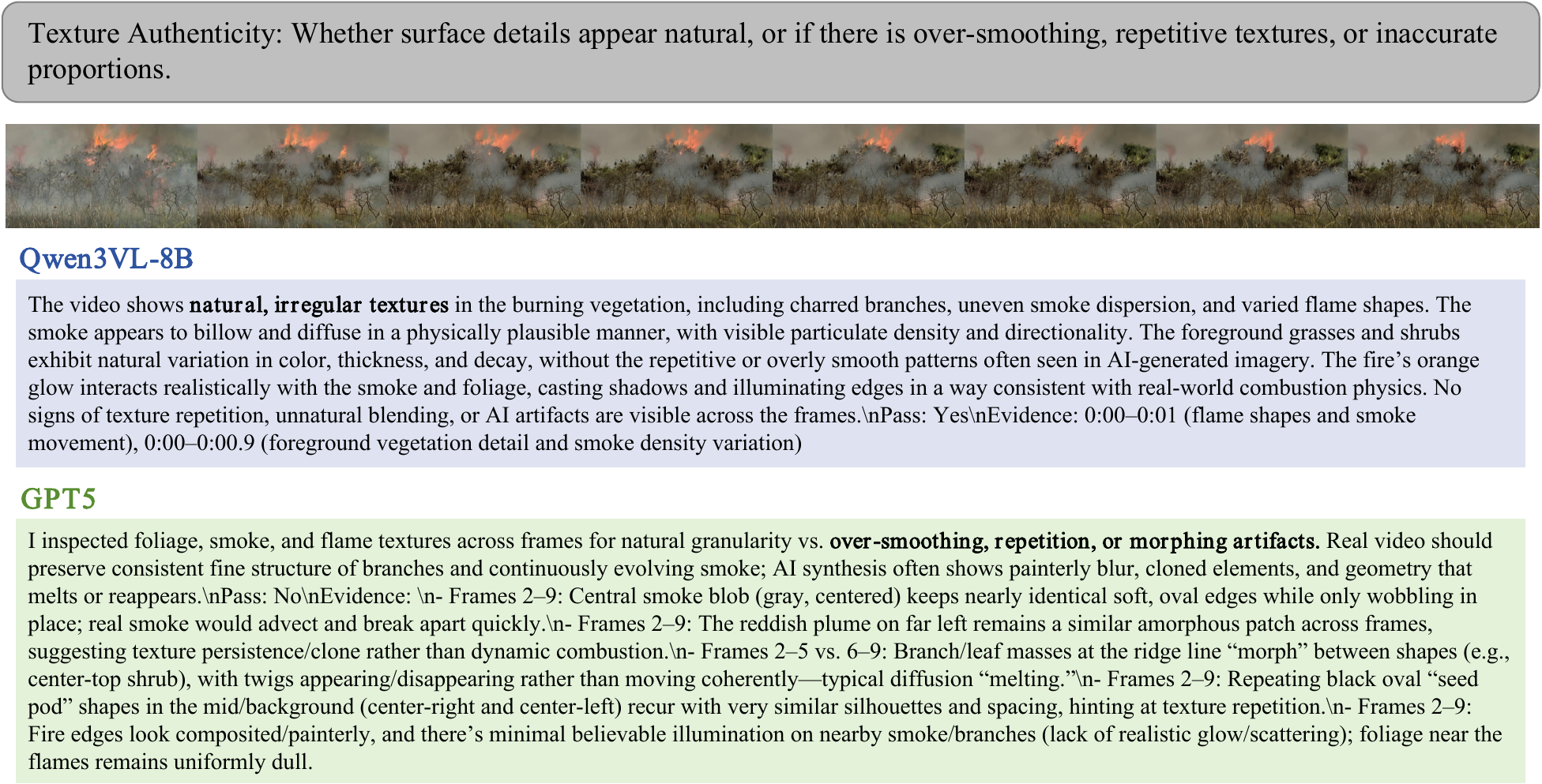}
  \caption{
    An illustrative example of Texture Authenticity, comparing the reasoning outputs of Qwen3VL-8B and GPT-5.
  }
  \label{fig:example_1}
\end{figure}

\begin{figure}[ht]
  \centering
  \includegraphics[width=0.8\textwidth]{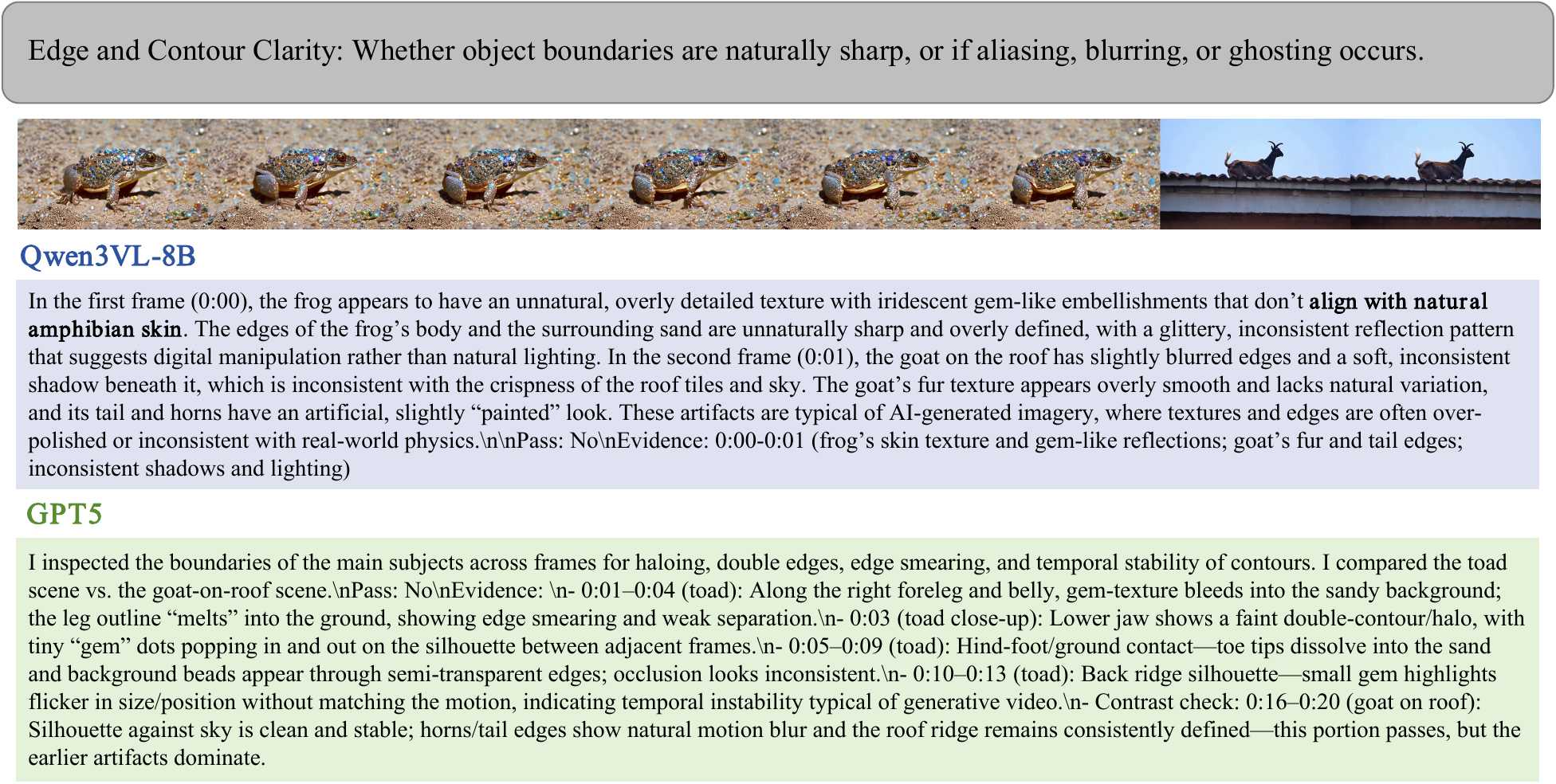}
  \caption{
    An illustrative example of Edge and Contour Clarity, comparing the reasoning outputs of Qwen3VL-8B and GPT-5.
  }
  \label{fig:example_2}
\end{figure}

\begin{figure*}[t]
  \centering
  \includegraphics[width=0.8\textwidth]{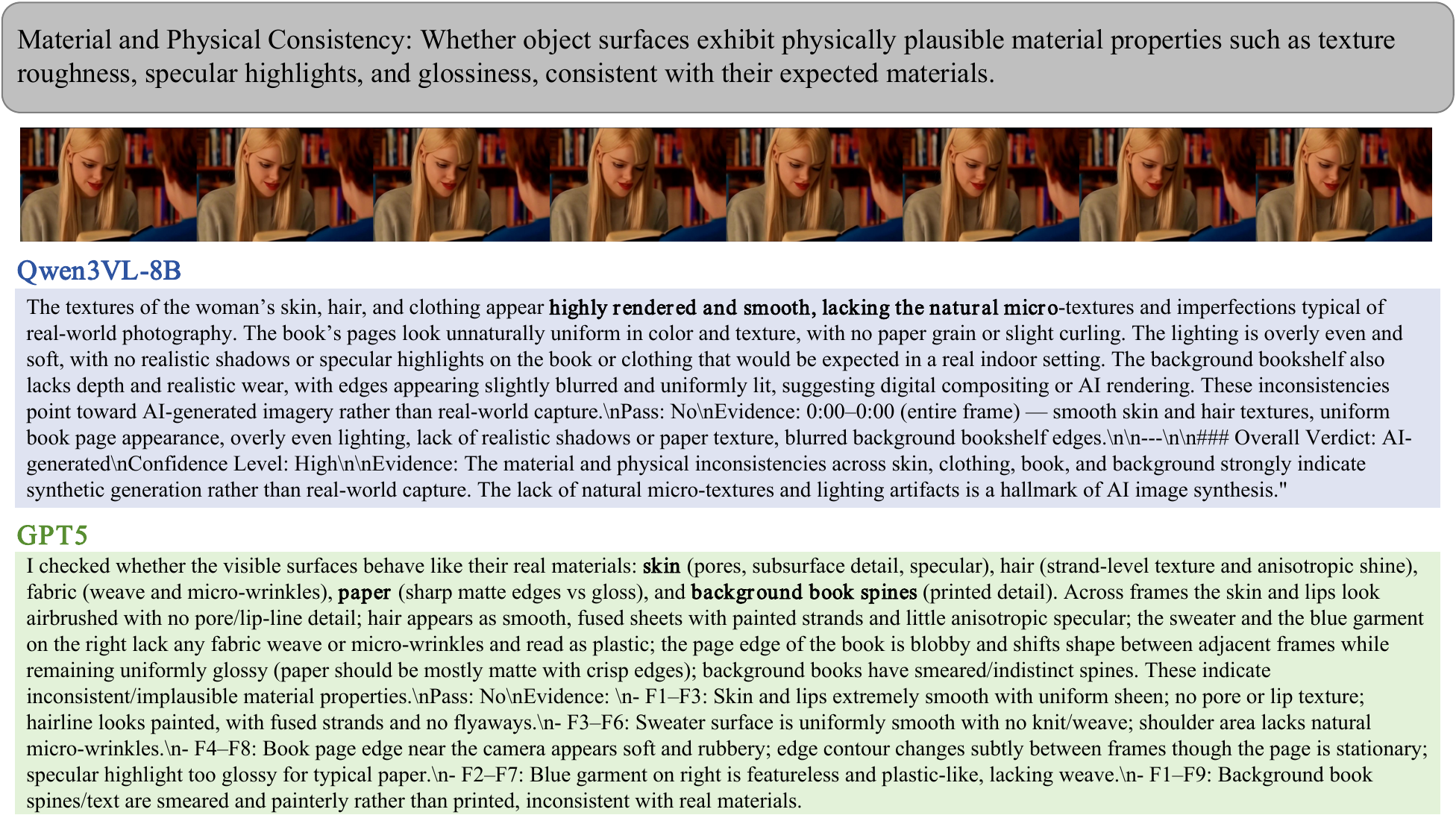}
  \caption{
   An illustrative example of Material and Physical Consistency, comparing the reasoning outputs of Qwen3VL-8B and GPT-5.
  }
  \label{fig:example_3}
\end{figure*}

\begin{figure*}[t]
  \centering
  \includegraphics[width=0.8\textwidth]{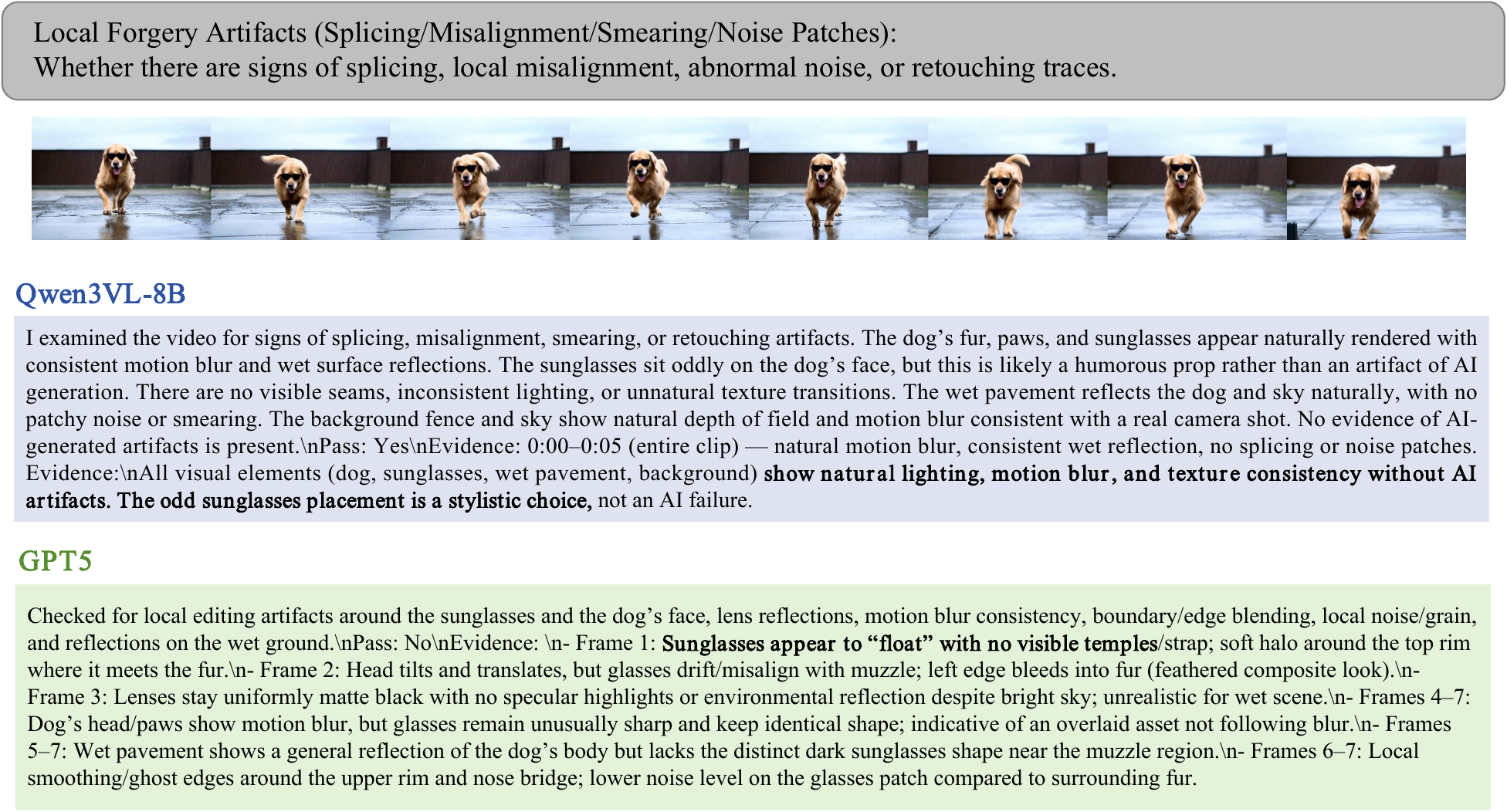}
  \caption{
    An illustrative example of Local Forgery Artifacts, comparing the reasoning outputs of Qwen3VL-8B and GPT-5.
  }
  \label{fig:example_4}
\end{figure*}

\begin{figure*}[t]
  \centering
  \includegraphics[width=0.8\textwidth]{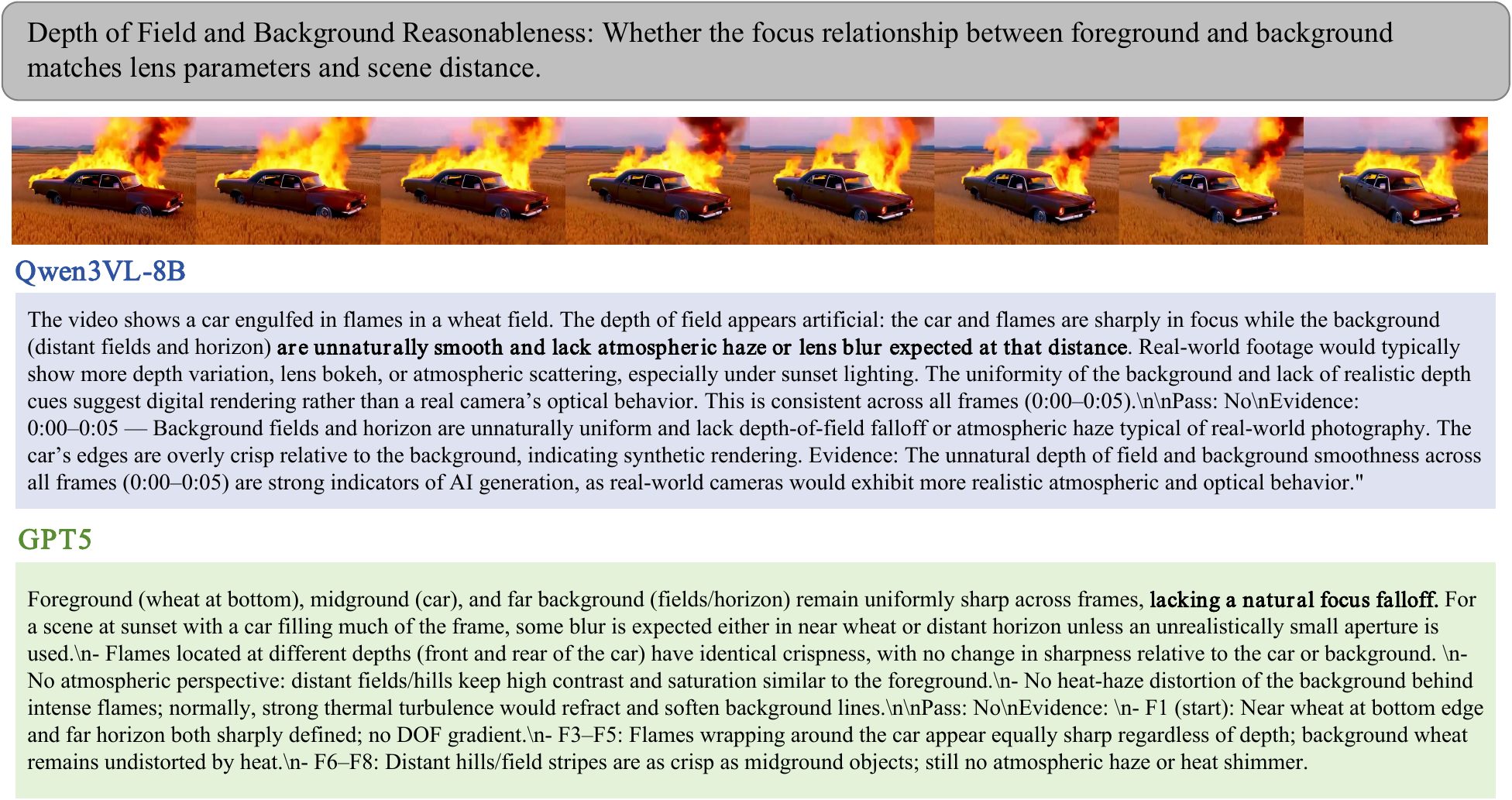}
  \caption{
   An illustrative example of Depth of Field and Background Reasonableness, comparing the reasoning outputs of Qwen3VL-8B and GPT-5.
  }
  \label{fig:example_5}
\end{figure*}

\begin{figure*}[t]
  \centering
  \includegraphics[width=0.8\textwidth]{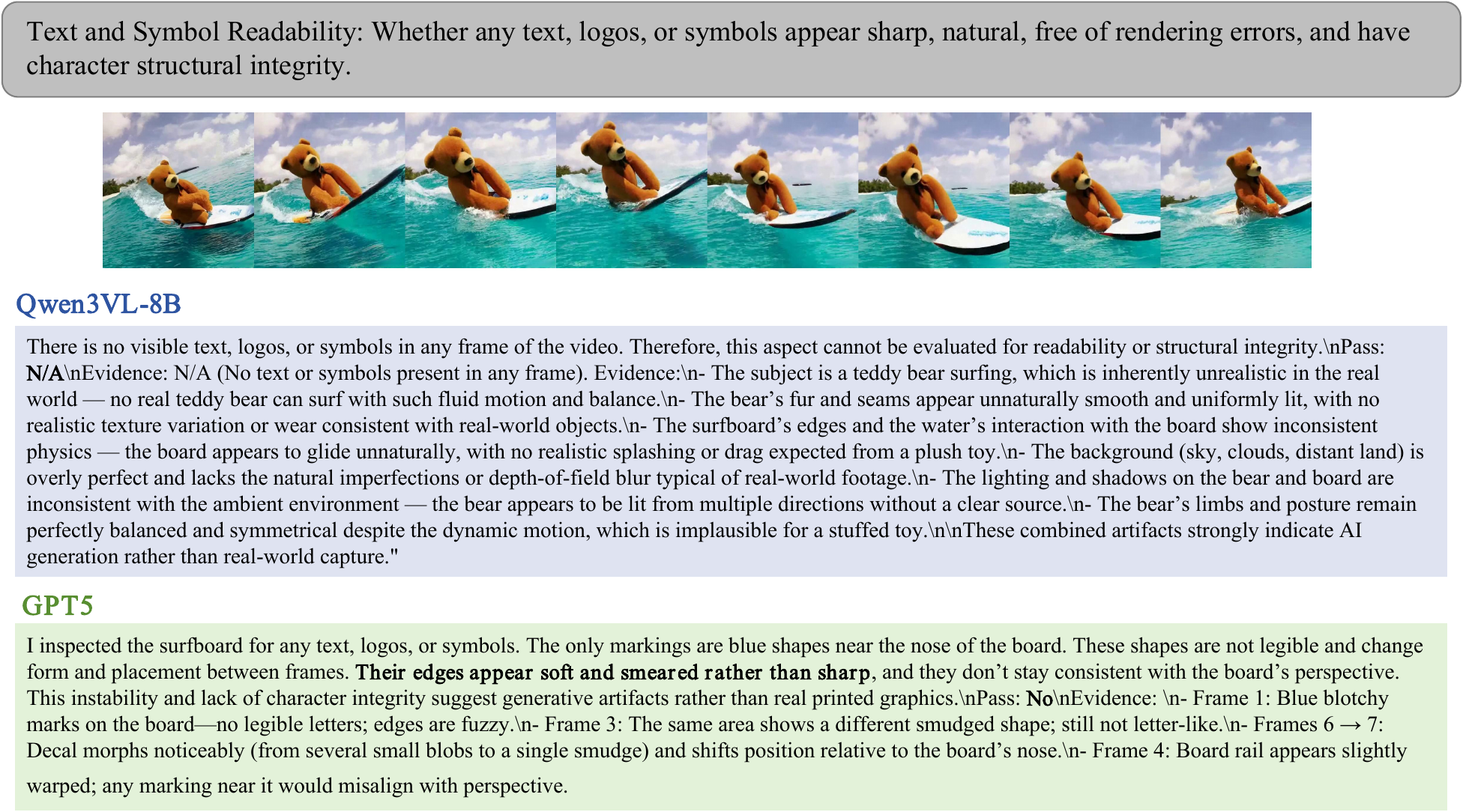}
  \caption{
    An illustrative example of Text and Symbol Readability, comparing the reasoning outputs of Qwen3VL-8B and GPT-5.
  }
  \label{fig:example_6}
\end{figure*}

\begin{figure*}[t]
  \centering
  \includegraphics[width=0.8\textwidth]{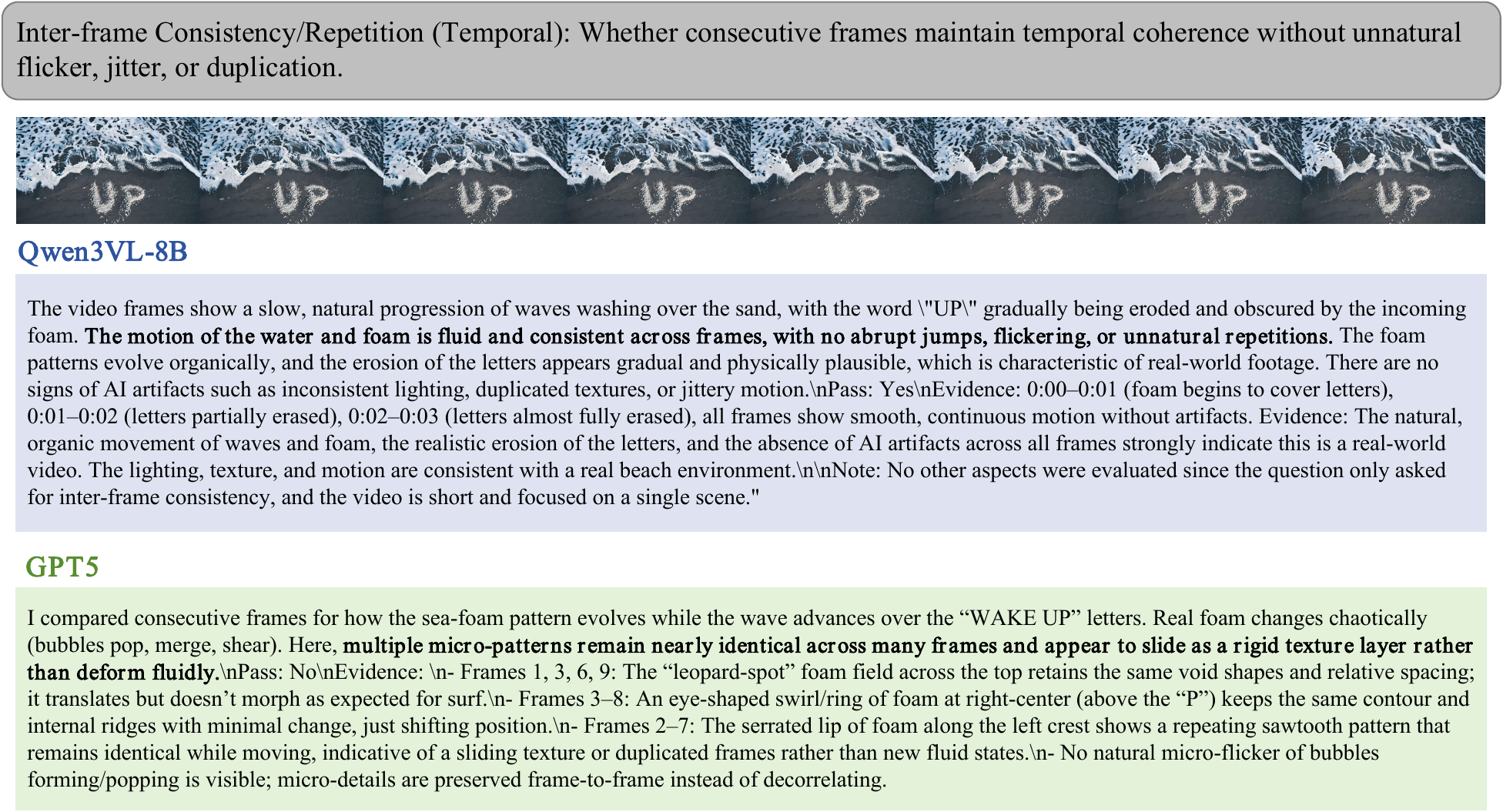}
  \caption{
     An illustrative example of Inter-frame Consistency/Repetition, comparing the reasoning outputs of Qwen3VL-8B and GPT-5.
  }
  \label{fig:example_7}
\end{figure*}

\begin{figure*}[t]
  \centering
  \includegraphics[width=0.8\textwidth]{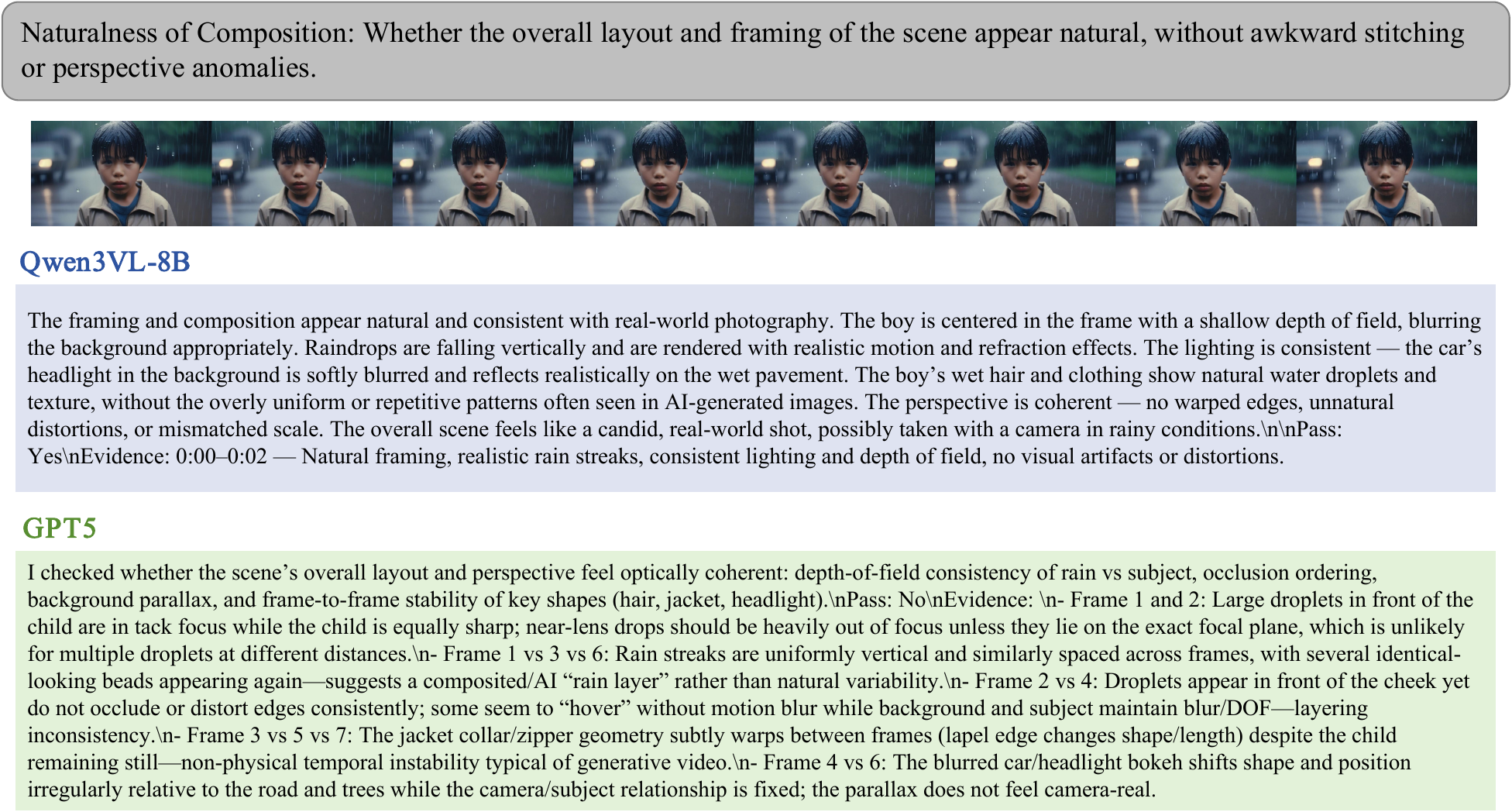}
  \caption{
    An illustrative example of Naturalness of Composition, comparing the reasoning outputs of Qwen3VL-8B and GPT-5.
  }
  \label{fig:example_8}
\end{figure*}

\begin{figure*}[t]
  \centering
  \includegraphics[width=0.8\textwidth]{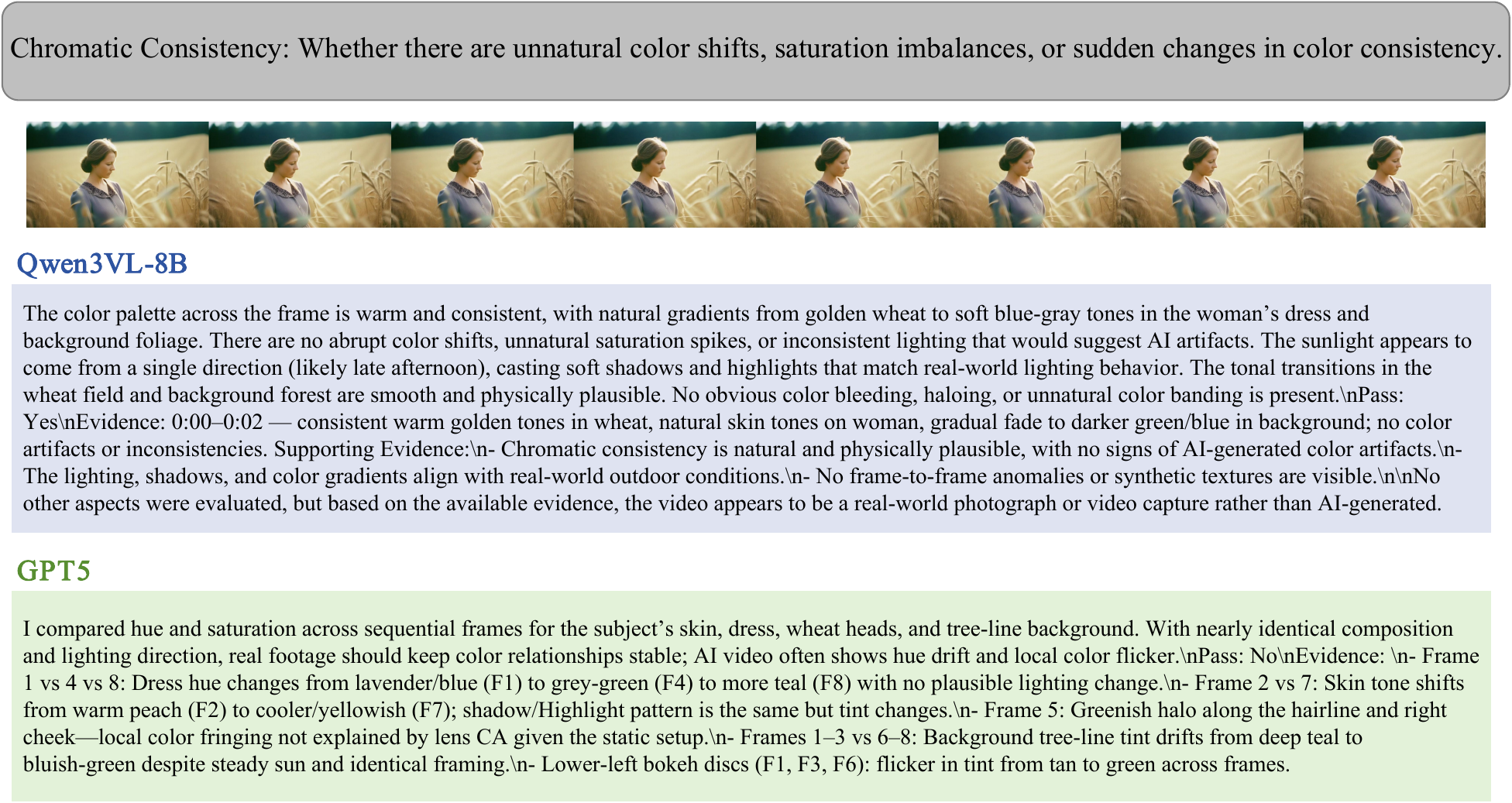}
  \caption{
    An illustrative example of Chromatic Consistency, comparing the reasoning outputs of Qwen3VL-8B and GPT-5.
  }
  \label{fig:example_9}
\end{figure*}

\begin{figure*}[t]
  \centering
  \includegraphics[width=0.8\textwidth]{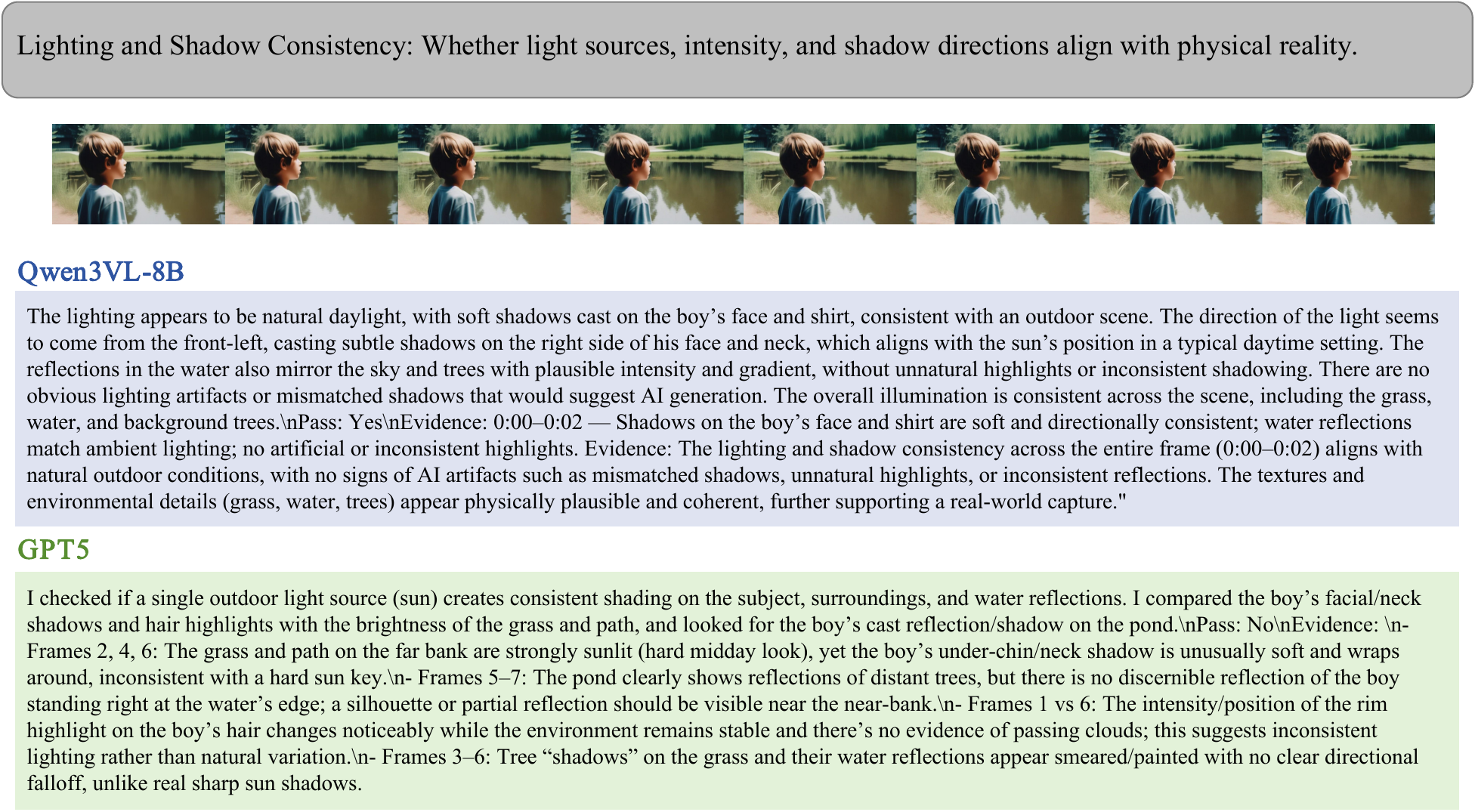}
  \caption{
    An illustrative example of Lighting and Shadow Consistency, comparing the reasoning outputs of Qwen3VL-8B and GPT-5.
  }
  \label{fig:example_10}
\end{figure*}

\begin{figure*}[t]
  \centering
  \includegraphics[width=0.8\textwidth]{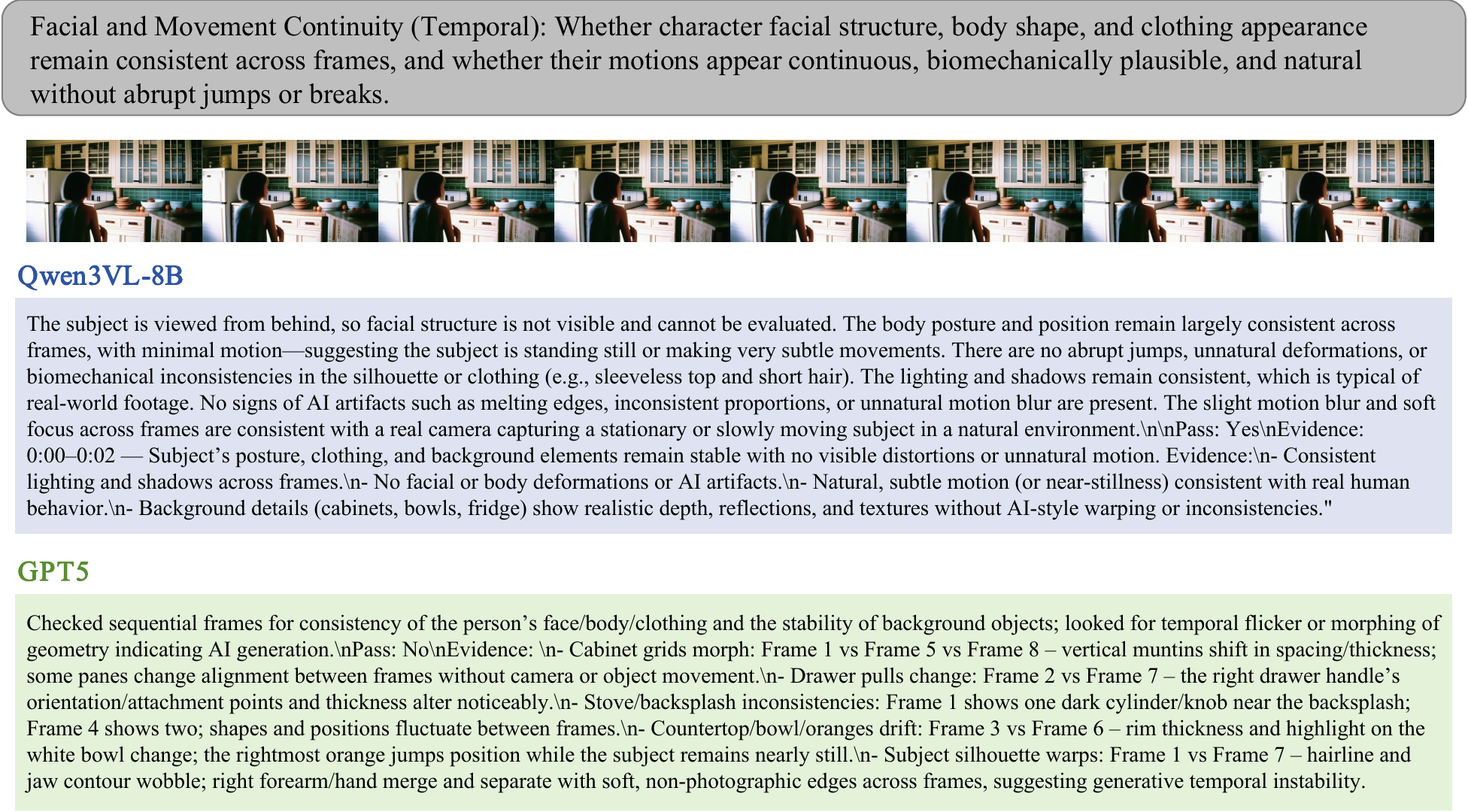}
  \caption{
    An illustrative example of Facial and Movement Continuity, comparing the reasoning outputs of Qwen3VL-8B and GPT-5.
  }
  \label{fig:example_11}
\end{figure*}

\begin{figure*}[t]
  \centering
  \includegraphics[width=0.8\textwidth]{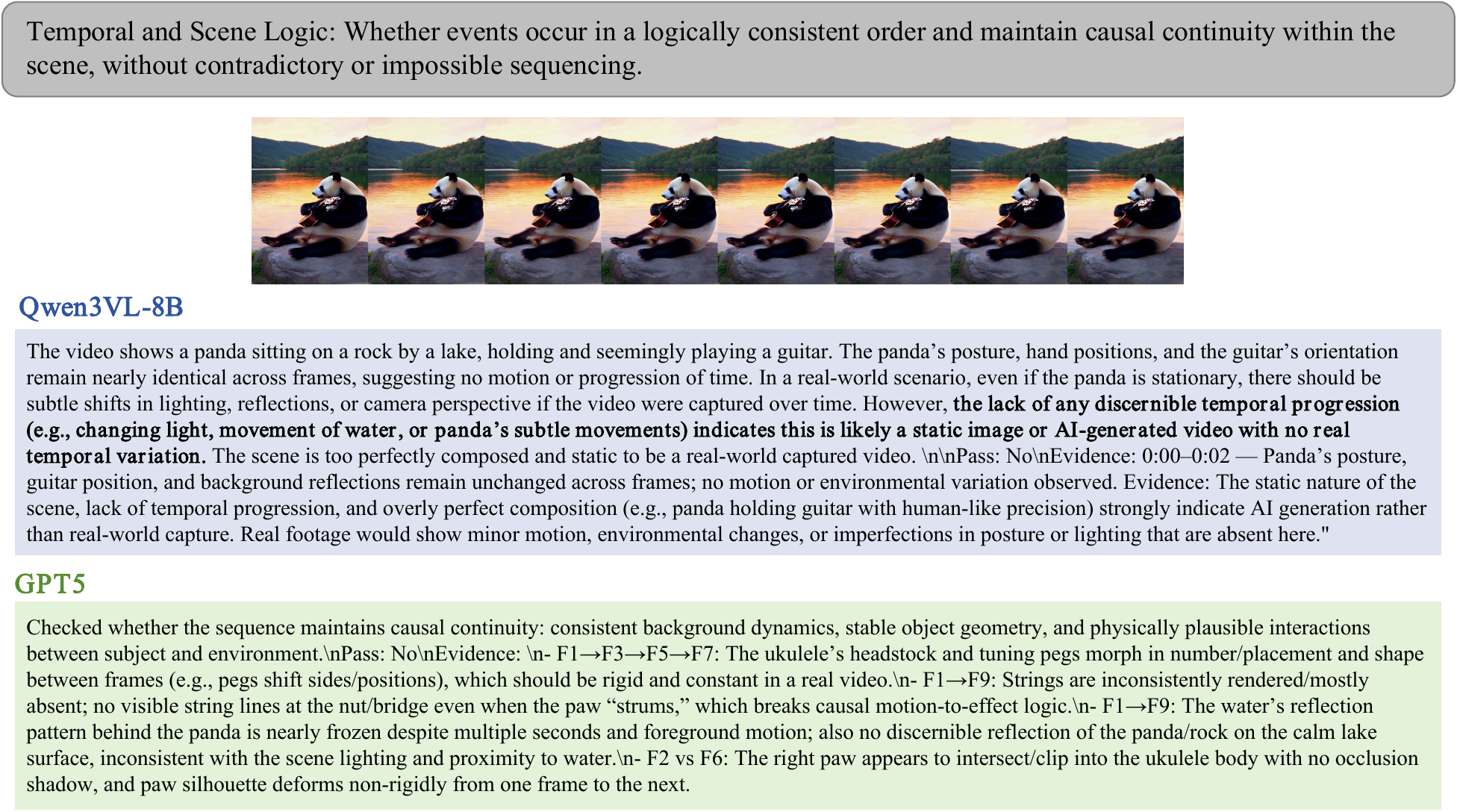}
  \caption{
        An illustrative example of Temporal and Scene Logic, comparing the reasoning outputs of Qwen3VL-8B and GPT-5.
  }
  \label{fig:example_12}
\end{figure*}

\begin{figure*}[t]
  \centering
  \includegraphics[width=0.8\textwidth]{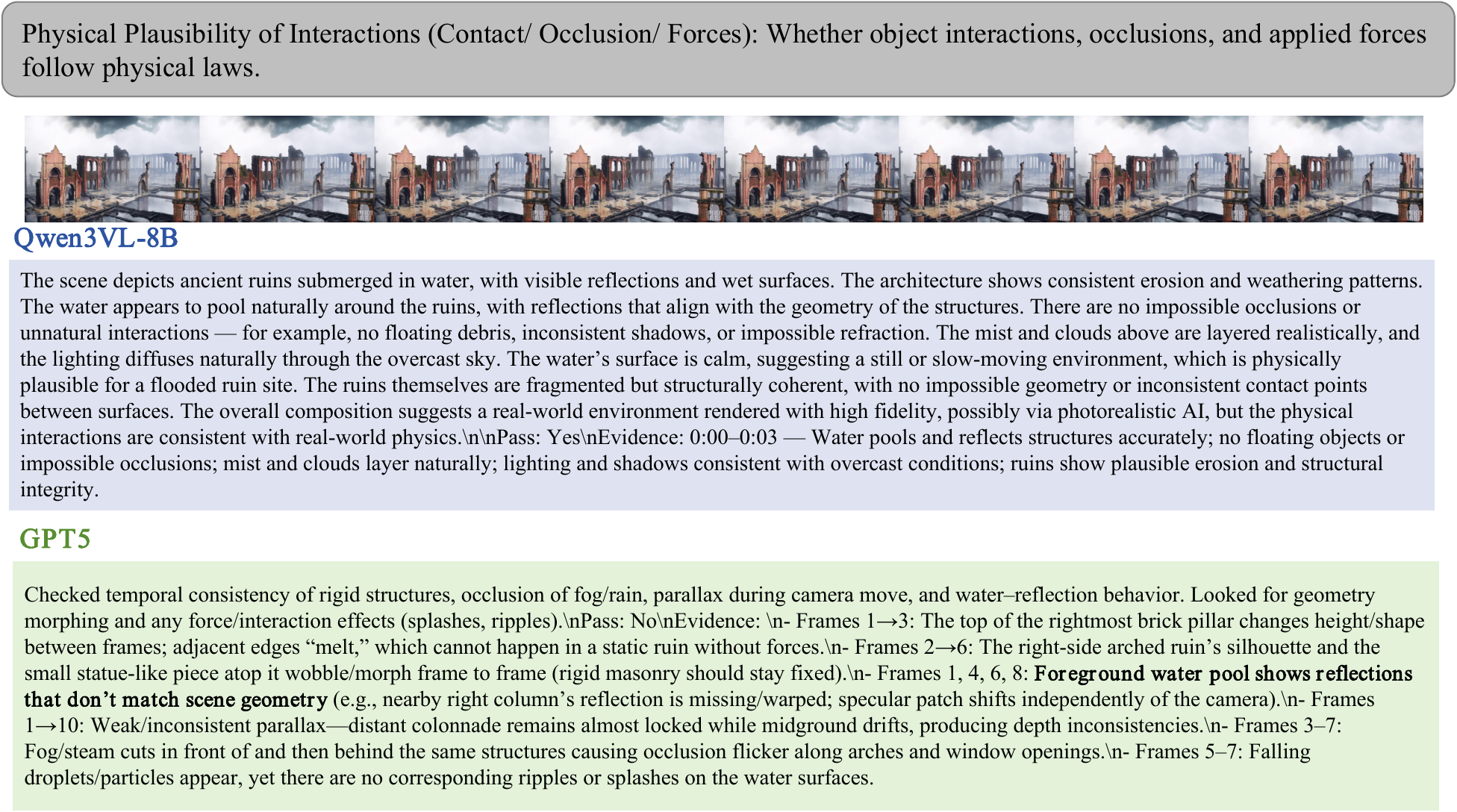}
  \caption{
       An illustrative example of Physical Plausibility of Interactions, comparing the reasoning outputs of Qwen3VL-8B and GPT-5.
  }
  \label{fig:example_13}
\end{figure*}

\begin{figure*}[t]
  \centering
  \includegraphics[width=0.8\textwidth]{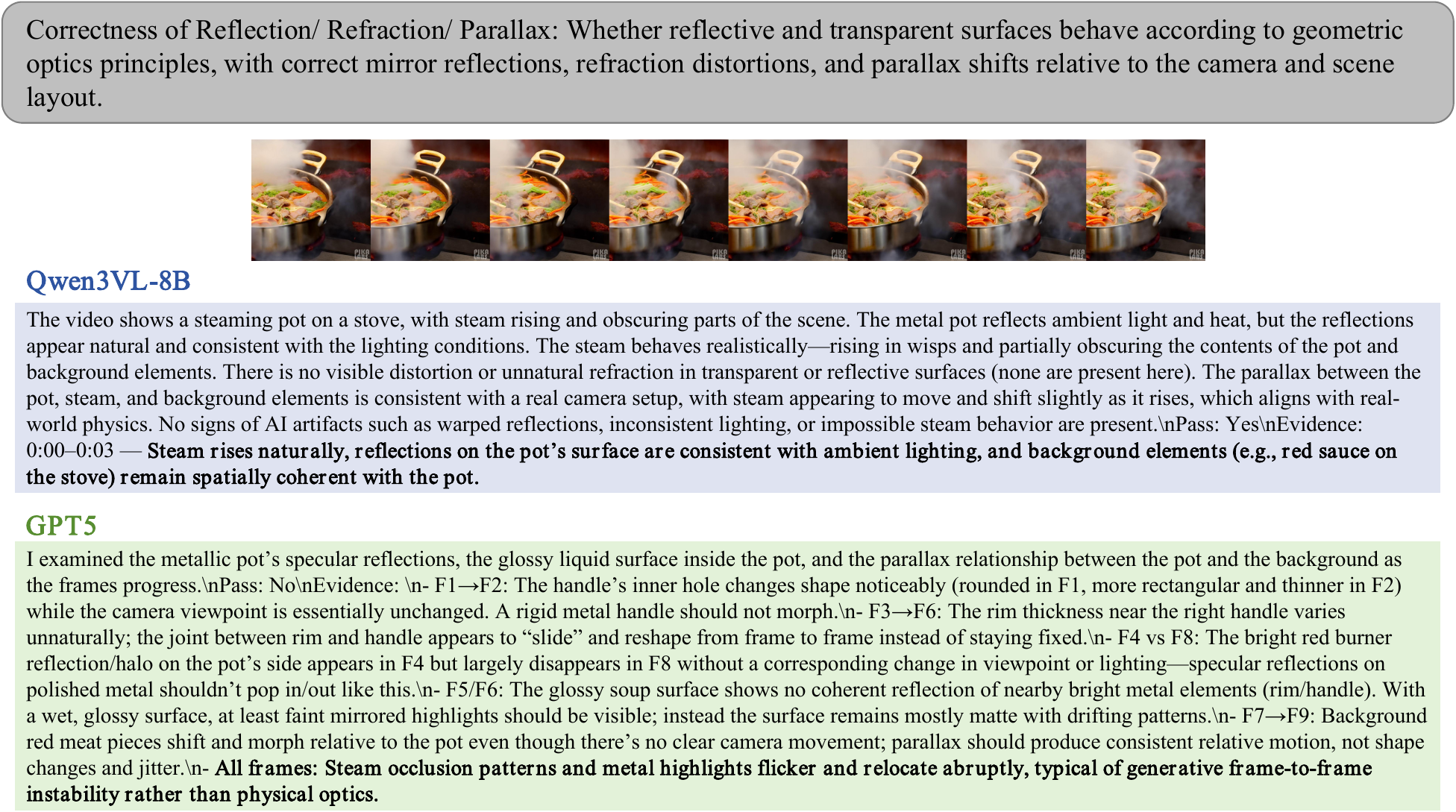}
  \caption{
        An illustrative example of Correctness of Reflection, comparing the reasoning outputs of Qwen3VL-8B and GPT-5.
  }
  \label{fig:example_14}
\end{figure*}

\begin{figure*}[t]
  \centering
  \includegraphics[width=0.8\textwidth]{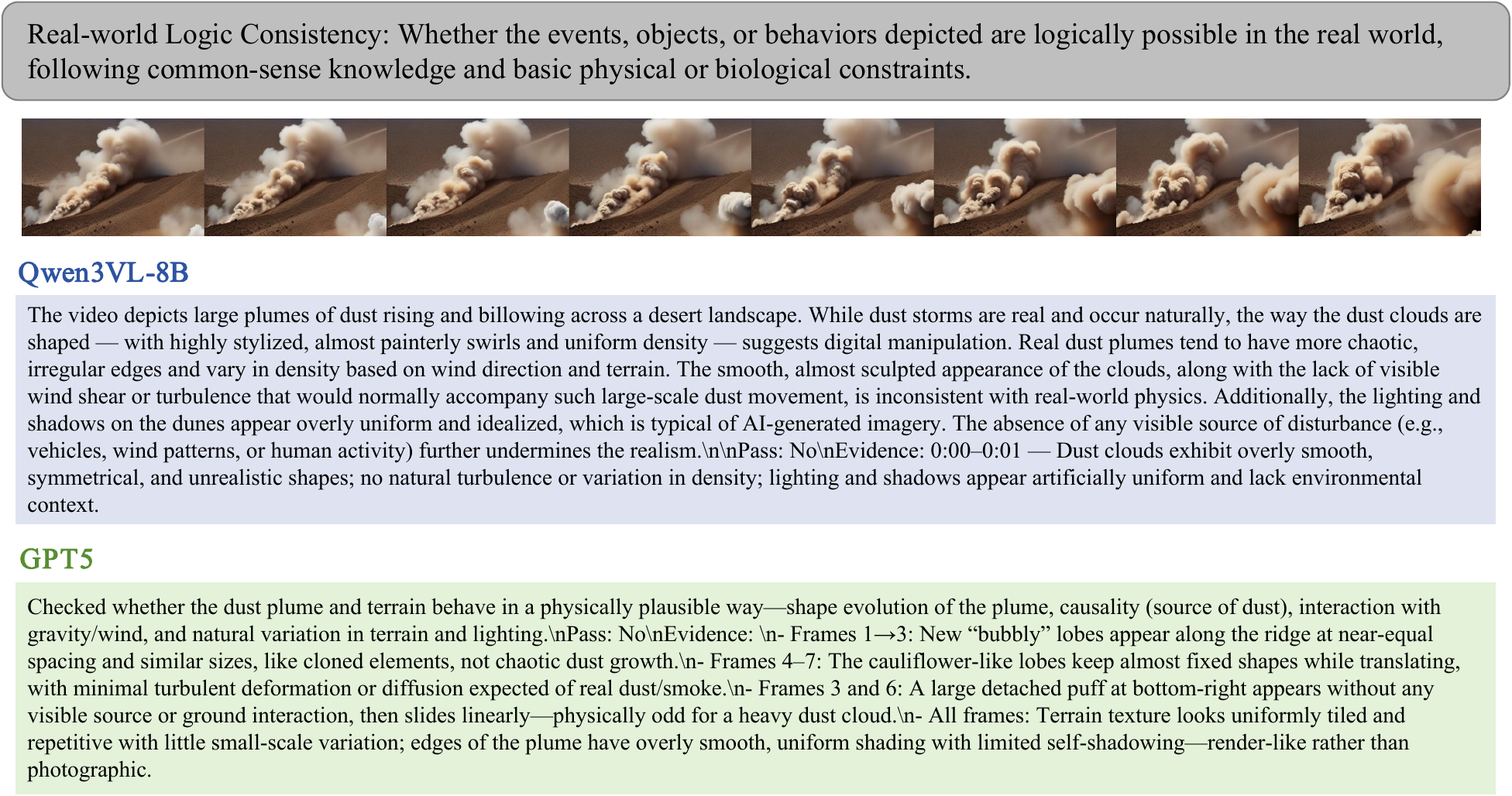}
  \caption{
        An illustrative example of Real-world Consistency, comparing the reasoning outputs of Qwen3VL-8B and GPT-5.
  }
  \label{fig:example_15}
\end{figure*}

For completeness and to facilitate transparent reproducibility, we provide in the following pages the example of 15-dimensional evaluation outputs generated by Qwen3-VL-8B and GPT-5 (see Figs.~\ref{fig:example_1}, \ref{fig:example_2}, \ref{fig:example_3},
\ref{fig:example_4}, \ref{fig:example_5}, \ref{fig:example_6},
\ref{fig:example_7}, \ref{fig:example_8}, \ref{fig:example_9}, \ref{fig:example_10}, \ref{fig:example_11}, 
\ref{fig:example_12}, \ref{fig:example_13}, \ref{fig:example_14}, 
\ref{fig:example_15}). 
These outputs include each model’s aspect-wise reasoning traces, intermediate evidence, and final decisions, allowing reviewers to directly inspect how the models behave across all diagnostic dimensions.

% \clearpage
% \section{Annotation Website}
% We also include visual illustrations of our annotation interface and the final conflict-bucket arbitration interface, enabling reviewers to clearly understand how annotations were collected, verified, and resolved.
% % \begin{figure}[ht]
% %   \centering
% %   \includegraphics[width=\textwidth]{sec/fig_appendix/expert_annotation_2.pdf}
% %   \caption{
% %        \textit{Where} Bench Video Annotation Website.
% %   }
% %   \label{fig:annotation}
% % \end{figure}

% % \begin{figure}[ht]
% %   \centering
% %   \includegraphics[width=\textwidth]{sec/fig_appendix/expert_annotation.pdf}
% %   \caption{
% %          Final review interface for conflict-bucket videos on the \textit{Where} Bench annotation website
% %   }
% %   \label{fig:annotation_2}
% % \end{figure}

% % % 
% % % 
% % To split the supplementary pages from the main paper, you can use \href{https://support.apple.com/en-ca/guide/preview/prvw11793/mac#:~:text=Delete%20a%20page%20from%20a,or%20choose%20Edit%20%3E%20Delete).}{Preview (on macOS)}, \href{https://www.adobe.com/acrobat/how-to/delete-pages-from-pdf.html#:~:text=Choose%20%E2%80%9CTools%E2%80%9D%20%3E%20%E2%80%9COrganize,or%20pages%20from%20the%20file.}{Adobe Acrobat} (on all OSs), as well as \href{https://superuser.com/questions/517986/is-it-possible-to-delete-some-pages-of-a-pdf-document}{command line tools}.

\end{document}